# Disentangled Representations for Short-Term and Long-Term Person Re-Identification

Chanho Eom ⑩, Wonkyung Lee ⑩, Geon Lee ⑩, and Bumsub Ham ⑩, *Member, IEEE*

**Abstract**—We address the problem of person re-identification (reID), that is, retrieving person images from a large dataset, given a query image of the person of interest. A key challenge is to learn person representations robust to intra-class variations, as different persons could have the same attribute, and persons' appearances look different, e.g., with viewpoint changes. Recent reID methods focus on learning person features discriminative only for a particular factor of variations (e.g., human pose), which also requires corresponding supervisory signals (e.g., pose annotations). To tackle this problem, we propose to factorize person images into identity-related and -unrelated features. Identity-related features contain information useful for specifying a particular person (e.g., clothing), while identity-unrelated features hold other factors (e.g., human pose). To this end, we propose a new generative adversarial network, dubbed *identity shuffle GAN* (IS-GAN). It disentangles identity-related and -unrelated features from person images through an identity-shuffling technique that exploits identification labels alone without any auxiliary supervisory signals. We restrict the distribution of identity-unrelated features, or encourage the identity-related and -unrelated features to be uncorrelated, facilitating the disentanglement process. Experimental results validate the effectiveness of IS-GAN, showing state-of-the-art performance on standard reID benchmarks, including Market-1501, CUHK03 and DukeMTMC-reID. We further demonstrate the advantages of disentangling person representations on a long-term reID task, setting a new state of the art on a Celeb-reID dataset. Our code and models are available online: https://cvlab-yonsei.github.io/projects/ISGAN/.

✦

## 1 INTRODUCTION

PERSON re-identification (reID) aims at retrieving person images of the same identity as a query from a large dataset, which is particularly important for finding/tracking missing persons or criminals in a surveillance system. This can be thought of as a *fine-grained* retrieval task in that 1) all images in the dataset belong to the same object class (i.e., person) with large intra-class variations (e.g., pose and scale changes), and 2) they are typically captured with different illumination conditions and background clutter across multiple cameras possibly with different characteristics and viewpoints. To tackle these problems, reID methods have focused on learning metric space [1], [2], [3], [4], [5], [6] and discriminative person representations [7], [8], [9], [10], [11], [12], [13], [14], [15], robust to intra-class variations and distracting scene details.

Convolutional neural networks (CNNs) have allowed significant advances in person reID in the past few years. ReID methods using CNNs add few more layers for aggregating

• *The authors are with the School of Electrical and Electronic Engineering, Yonsei University, Seoul 03722, South Korea. E-mail: {cheom, wonkyung.lee, geon.lee, bumsub.ham}@yonsei.ac.kr.*

*Manuscript received 16 December 2020; revised 2 July 2021; accepted 16 October 2021. Date of publication 26 October 2021; date of current version 3 November 2022.*
*This research was supported by R&D program for Advanced Integrated-intelligence for Identification (AIID) through the National Research Foundation of KOREA(NRF) funded by Ministry of Science and ICT under Grant NRF-2018M3E3A1057289, and Yonsei University Research Fund of 2021 (2021-22-0001).*
*(Corresponding author: Bumsub Ham.)*
*Recommended for acceptance by T. Xiang.*
*Digital Object Identifier no. 10.1109/TPAMI.2021.3122444*

body parts [9], [10], [11], [12], [16], [17] and/or computing attention maps [13], [14], [15], on the top of, e.g., a (cropped) ResNet [18] trained for ImageNet classification [19]. They give state-of-the-art results, but finding person representations robust to various factors is still very challenging. Recent methods exploit generative adversarial networks (GANs) [20] to learn feature representations robust to a particular factor. For example, conditioned on a target pose map and a person image, they generate a new person image of the same identity but having the target pose [21], [22], and the generated images are used as additional training samples. This allows to learn pose-invariant features, and also has an effect of data augmentation for regularization.

In this paper, we introduce a novel framework, dubbed *identity shuffle GAN* (IS-GAN), that disentangles identity-related/-unrelated features from input person images, without any auxiliary supervisory signals, except identification labels. Identity-related features contain information useful for identifying a particular person (e.g., gender and body shape), while identity-unrelated ones hold all other attributes (e.g., human pose, background clutter, occlusion, and scale changes). We visualize in Fig. 1 novel person images generated using identity-related/-unrelated features. Specifically, the person images are synthesized with new person representations obtained by interpolating either identity-related or -unrelated features between two person images, while fixing the other ones. We can see that the cloth colors are blended, while human pose and background clutter remain unchanged, when interpolating the identity-related features. On the other hand, when identity-unrelated features are altered, human pose and background clutter are mixed, but the cloth colors are maintained.

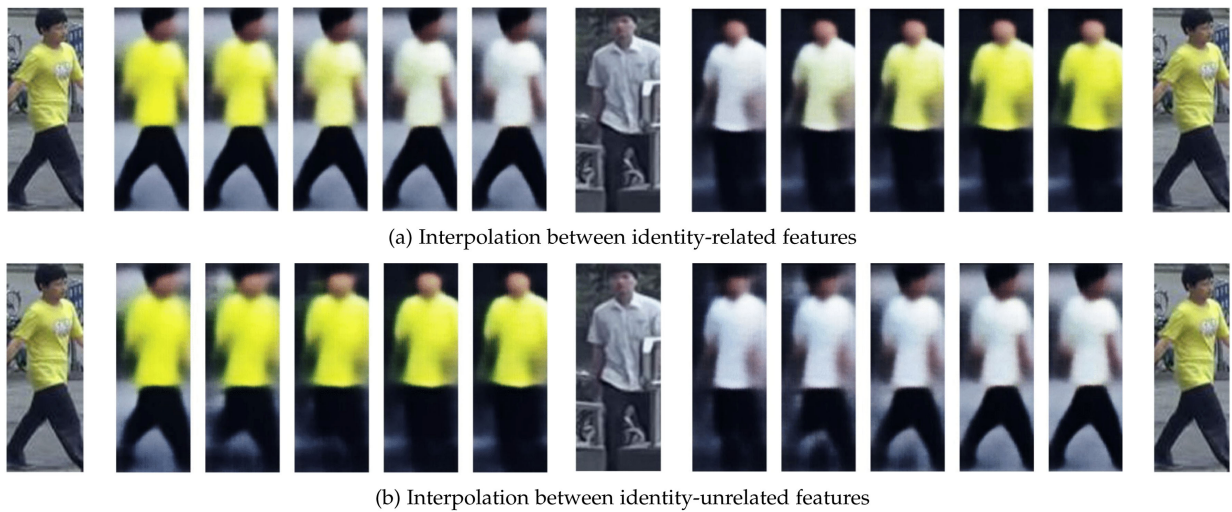

(a) Interpolation between identity-related features

(b) Interpolation between identity-unrelated features

Fig. 1. Visual comparison of identity-related and -unrelated features. We generate new person images by interpolating (a) identity-related features and (b) identity-unrelated ones between two images, while fixing the other ones. We can see that identity-related features encode, e.g., clothing and color, and identity-unrelated ones involve, e.g., human pose and background clutter. Note that we disentangle these features using identification labels only. (Best viewed in color.).

For the disentanglement process, we propose an identity shuffling technique that allows to factorize the identity-related/-unrelated features using identification labels alone. At training time, IS-GAN inputs person images of the same identity, and extracts identity-related/-unrelated features. In particular, we divide person images into horizontal parts, and disentangle these features in both image- and part-levels. We then learn to generate new images of the same identity by shuffling identity-related features between the person images. We further facilitate the disentanglement process by restricting the distribution of identity-unrelated features [23], [24], [25], [26], limiting the information they could have. They, however, may encode in part identity-related attributes, as the restriction does not guarantee that identity-related/-unrelated features are mutually exclusive. To address this, we introduce another regularization technique that minimizes correlations between these features explicitly, encouraging them to be mutually exclusive. We propose two models depending on which regularization technique is used, and analyze the effect of each technique within our framework. At test time, we use the identity-related features only to retrieve person images. We demonstrate the effectiveness of our approach in both short- and long-term reID tasks on standard benchmarks. Note that the long-term reID task is much more challenging than the short-term one, as it considers re-identifying persons after a long period of time, and thus an outfit of the person in a dataset could be different from that of the query person. We set new state of the art on both cases, and show an extensive experimental analysis with ablation studies. The main contributions of this paper are threefold as follows:

- We introduce a novel method for person reID, dubbed IS-GAN, which disentangles person images into identity-related/-unrelated features using an identity shuffling technique in both image- and part-levels without any auxiliary supervisory signals.
- We propose a decorrelation loss that encourages identity-related/-unrelated features to be mutually exclusive.

- We achieve the state of the art on both short- and long-term reID benchmarks, and demonstrate the effectiveness of our approach, with extensive experimental results and detailed analyses.

A preliminary version of this work appeared in [27]. This version adds 1) a detailed description of related works including long-term reID; 2) a regularization technique that removes correlations between identity-related and -unrelated features explicitly; 3) a detailed analysis on the effect of regularization techniques within our framework; 4) more comparisons with the state of the art on standard reID benchmarks, including Market-1501 [28], CUHK03 [29], and DukeMTMC-reID [30]; 5) quantitative and qualitative comparisons with the state of the art on a long-term reID task; 6) an in-depth analysis on what identity-related features encode; 7) visual analyses on identity-related/-unrelated features for both long- and short-term reID tasks; 8) detailed descriptions for the differences from DG-Net [31]; 9) a sensitivity analysis on hyperparameters of our model.

## 2 RELATED WORK

In this section, we briefly introduce representative works related to ours, including short-term and long-term reID methods and disentangled feature representations.

### 2.1 Short-Term reID

ReID methods typically provide person representations robust to a particular factor, such as human pose, occlusion, and background clutter. Part-based methods represent a person image as a combination of body parts either explicitly or implicitly [9], [10], [11], [12], [16], [17], [32], [33], [34]. Explicit part-based methods use off-the-shelf human parsing/pose estimation models, and extract body parts (e.g., head and torso) with corresponding features [9], [10], [33], [34]. This makes it possible to obtain pose-invariant representations, but the human parsing/pose estimation models often give incorrect pose maps, especially for occluded parts. Instead of using human pose explicitly, a person image is sliced into different horizontal parts of multiple scales in implicit part-based methods [11], [12], [17]. They exploit various local regions of the image, and provide feature representations robust to occlusion. Spatial attention techniques are also widely leveraged in person reID to localize discriminative body parts in images [13], [14], [15], providing person representations robust against background clutter. Channel attention can further be used to refine the representations with high-level semantic contexts [35], [36], [37].

Recent reID methods leverage GANs to fill the domain gap between source and target datasets [38], [39] or to obtain pose-invariant features [21], [22], [40]. In [38], CycleGAN [41] is used to transform person images from a source domain to a target one. Similarly, StarGAN [42] is exploited to match the camera styles between source and target domains [39]. Two typical ways of obtaining person representations robust to human pose are to fuse all features extracted from person images of different poses and to distill pose-relevant information. In [21], [22], new images are generated using GANs, conditioned on target pose maps and input person images. Person representations for the generated images are then fused, giving pose-invariant features. These approaches, however, require auxiliary pose information at test time. To address this problem, FD-GAN [40] proposes to distill identity-related and pose-unrelated features from person images. It generates new images of the same identity as an input but having target poses, enabling getting rid of pose-related information disturbing the reID task. Although FD-GAN does not require additional human pose information during inference, it offers limited feature representations, in that they are vulnerable to other factors of variations, such as scale changes, background clutter, and occlusion. Disentangling features with respect to these factors is infeasible within the FD-GAN framework, as it requires corresponding supervisory signals describing the factors (e.g., foreground masks for background clutter). In contrast, our model factorizes identity-related/-unrelated features without any auxiliary supervisory signals. We also propose to shuffle identity-related features in both image- and part-levels, which is helpful for extracting robust person representations, even from person images with occlusion and large pose variations. DG-Net [31] is the most similar work to ours in that it disentangles person representations into appearance and structure features. It uses a stylization technique [43] for the disentanglement, suggesting that the appearance and structure features encode style (e.g., color and texture) and content (e.g., gender and pose) information, respectively, rather than identity-related and -unrelated one. Furthermore, this method also does not consider a part-level feature consistency. DG-Net++ [44] further extends the idea of DG-Net, and applies it to an unsupervised domain adaptation problem for person reID. Similar to DG-Net, it leverages a stylization technique [43] for disentangling domain-invariant and -specific features.

## 2.2 Long-Term reID

Most reID methods [11], [12], [27], [31], [45] mainly focus on a short-term reID task, where a person of interest reappears under another camera in a short period of time (e.g., less than 30 minutes). They assume that the person of interest in a query and gallery images wear the same outfit. As this assumption may not hold after a long time span (e.g., one day after), directly employing typical reID methods [11], [12], [15], [32], [46], [47], e.g., exploiting clothes as discriminative features, for long-term reID is limited. ReIDCaps [48] proposes to use a capsule network [49] to model visual changes of each identity explicitly. While this approach shows competitive performance on long-tern reID, the number of appearance variations affordable to handle is limited by the dimension of capsules. Identity-shuffling in our approach enables eliminating features related to person outfits/belongings from identity-related ones, suggesting that it can be applied to the long-term reID task directly without any bells and whistles. It also does not restrict the number of appearance variations to handle.

New datasets for long-term reID have recently been introduced, such as Celeb-reID [48], COCAS [50], Div-Market [51], which consist of multiple images of the same identity but with different clothes and attributes. COCAS [50] and Div-Market use GANs to synthesize realistic images, since it is hard to track persons for a long time and to annotate identification labels manually. The generative models, however, cause inevitable artifacts. Celeb-reID [48] consists of street snap-shots of celebrities on the internet. It provides lots of images for persons wearing different clothes in various scenes, posing a challenge of a practical long-term reID scenario.

## 2.3 Disentangled Representations

Disentangling the factors of variations in CNN features has been widely leveraged for image synthesis [24], [52], image-to-image translation [25], [53], or feature distillation [23], [40], [54]. $D^2AE$ [52] and IP-GAN [24] disentangle identity and attribute features from face images for face identification and image synthesis, respectively. Similarly, MUNIT [53] and DRIT [25] decompose images into domain-specific and -invariant features for generating diverse outputs. The work of [26] leverages disentangled feature representations for domain-specific image deblurring. In [23], a conditional generative model is exploited to extract class-related and -independent features for image retrieval. Unlike these methods, DR-GAN [54] and FD-GAN [40] use a side information (i.e., pose labels) to learn identity-related and pose-unrelated features for face recognition and person reID, respectively.

# 3 APPROACH

In this section, we describe an overview of our framework, and explain our baseline model. We then present a detailed description of IS-GAN disentangling identity-related/-unrelated features for person reID.

## 3.1 Overview

We denote by $\mathbf{I}$ and $y \in \{1, 2, \ldots, C\}$ a person image and an identification label, respectively, where $C$ is the number of identities in a dataset. We denote by $\mathbf{I}_a$ and $\mathbf{I}_p$ anchor and positive images, respectively, that share the same identification label. At training time, we input pairs of $\mathbf{I}_a$ and $\mathbf{I}_p$ with corresponding identification labels, and train our model to learn identity-related/-unrelated features, denoted by $\phi_R(\mathbf{I})$ and $\phi_U(\mathbf{I})$, respectively. At test time, we compute

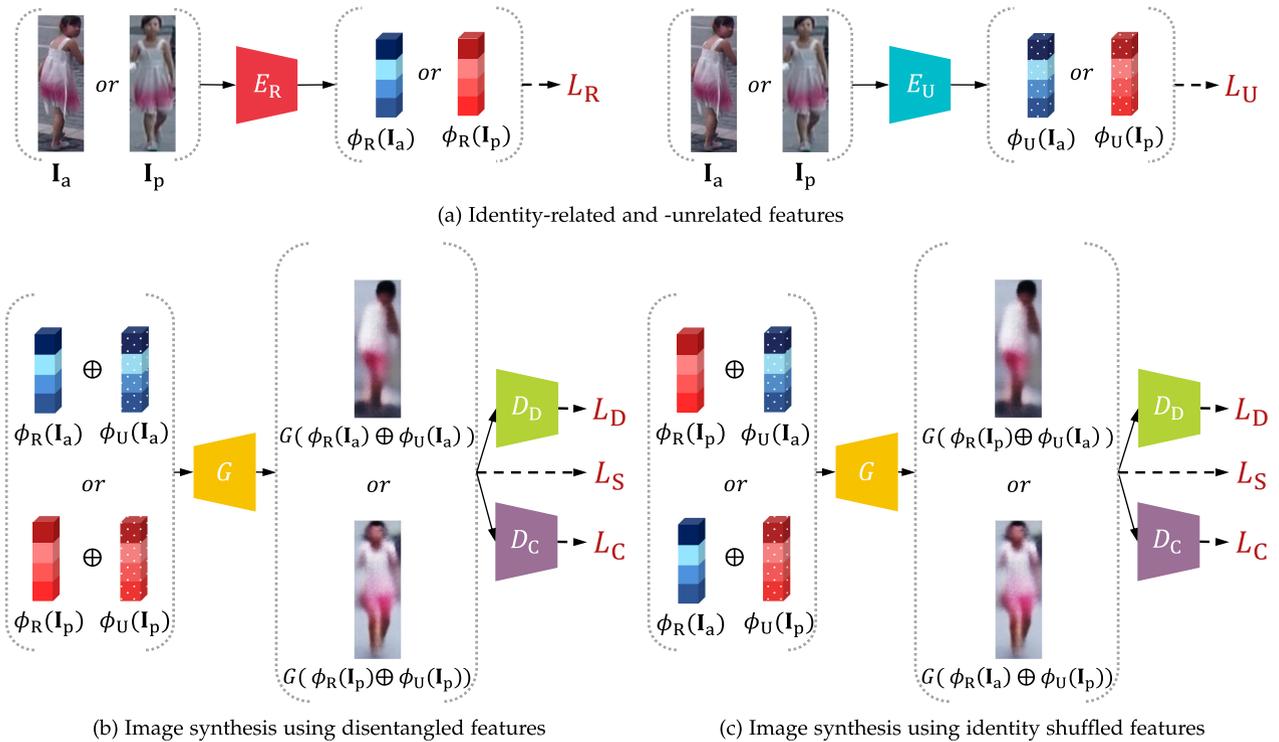

(a) Identity-related and -unrelated features

(b) Image synthesis using disentangled features

(c) Image synthesis using identity shuffled features

Fig. 2. Overview of IS-GAN. (a) IS-GAN disentangles identity-related/-unrelated features from input person images, i.e., anchor and positive images, denoted by $\mathbf{I}_a$ and $\mathbf{I}_p$, respectively. (b-c) To regularize the disentanglement process, it learns to generate the same images as the inputs using disentangled features (b) without and (c) with identity shuffling. The identity shuffling technique encourages 1) the identity-related encoder $E_R$ to extract the shared information between $\mathbf{I}_a$ and $\mathbf{I}_p$, and 2) the identity-unrelated encoder $E_U$ to see other factors in each image. We train all components of our model end-to-end, including the encoders ($E_R$ and $E_U$), the generator ($G$), and the discriminators ($D_D$ and $D_C$). We denote by $\oplus$ an element-wise addition of features. See text for details. (Best viewed in color.)

the euclidean distances between identity-related features of person images to determine whether the identities are the same or not.

IS-GAN mainly consists of five components (Fig. 2): An identity-related encoder $E_R$, an identity-unrelated encoder $E_U$, a generator $G$, a domain discriminator $D_D$, and a class discriminator $D_C$. Given pairs of $\mathbf{I}_a$ and $\mathbf{I}_p$, the encoders, $E_R$ and $E_U$, learn identity-related features, $\phi_R(\mathbf{I}_a)$ and $\phi_R(\mathbf{I}_p)$, and identity-unrelated ones, $\phi_U(\mathbf{I}_a)$ and $\phi_U(\mathbf{I}_p)$, respectively (Fig. 2a). To encourage identity-related and -unrelated encoders to disentangle these features from the input images, we train the generator $G$, such that it synthesizes the same images as $\mathbf{I}_a$ from $\phi_R(\mathbf{I}_a) \oplus \phi_U(\mathbf{I}_a)$ and $\phi_R(\mathbf{I}_p) \oplus \phi_U(\mathbf{I}_a)$, where we denote by $\oplus$ an element-wise addition of features (Figs. 2b and 2c). Similarly, it generates the same images as $\mathbf{I}_p$ from $\phi_R(\mathbf{I}_p) \oplus \phi_U(\mathbf{I}_p)$ and $\phi_R(\mathbf{I}_a) \oplus \phi_U(\mathbf{I}_p)$. Since $\mathbf{I}_a$ and $\mathbf{I}_p$ share the same identity but with, e.g.,; different poses, scales, and illumination, this identity shuffling allows the identity-related encoder $E_R$ to extract features robust to such factors, focusing on the shared attributes between $\mathbf{I}_a$ and $\mathbf{I}_p$, while encouraging the identity-unrelated encoder $E_U$ to capture others. We also perform the feature disentanglement and identity shuffling in a part-level by dividing the input images into multiple horizontal regions (Fig. 3). Given the generated images, the class discriminator $D_C$ determines their identities as either those for $\mathbf{I}_a$ or $\mathbf{I}_p$, and the domain discriminator $D_D$ tries to distinguish real and fake images. IS-GAN is trained end-to-end using identification labels alone without any auxiliary supervisory signals.

### 3.2 Baseline Model

To the baseline, we use the identity-related encoder $E_R$ only, trained with an identity-related loss. For the encoder $E_R$, we exploit a network architecture, similar to [12]. It has three branches on top of a backbone network, where each branch has the same structure but different parameters. We call them part-1, part-2, and part-3 branches, that slice a feature map from the backbone network equally into one, two, and three horizontal regions, respectively. The part-1 branch provides a global feature of the entire person image. Other branches give both global and local features describing body parts, where the local features are extracted from corresponding horizontal regions. For example, the part-3 branch outputs three local features and a single global one. Accordingly, we extract $K$ features from the encoder $E_R$ in total, where $K = 8$ in our case. Without loss of generality, we can use additional branches to consider different horizontal regions of multiple scales.

#### 3.2.1 Identity-Related Loss

We denote by $\mathbf{I}^k$ and $\phi_R^k$ $(k = 1, \ldots, K)$ horizontal regions of multiple scales and corresponding embedding functions that extract identity-related features, respectively. Following other reID methods [11], [12], [15], [22], we formulate the reID problem as a multi-class classification task, and train the encoder $E_R$ with a cross-entropy loss. Concretely, a loss function $\mathcal{L}_R$ to learn the embedding function $\phi_R^k$ is defined as follows:

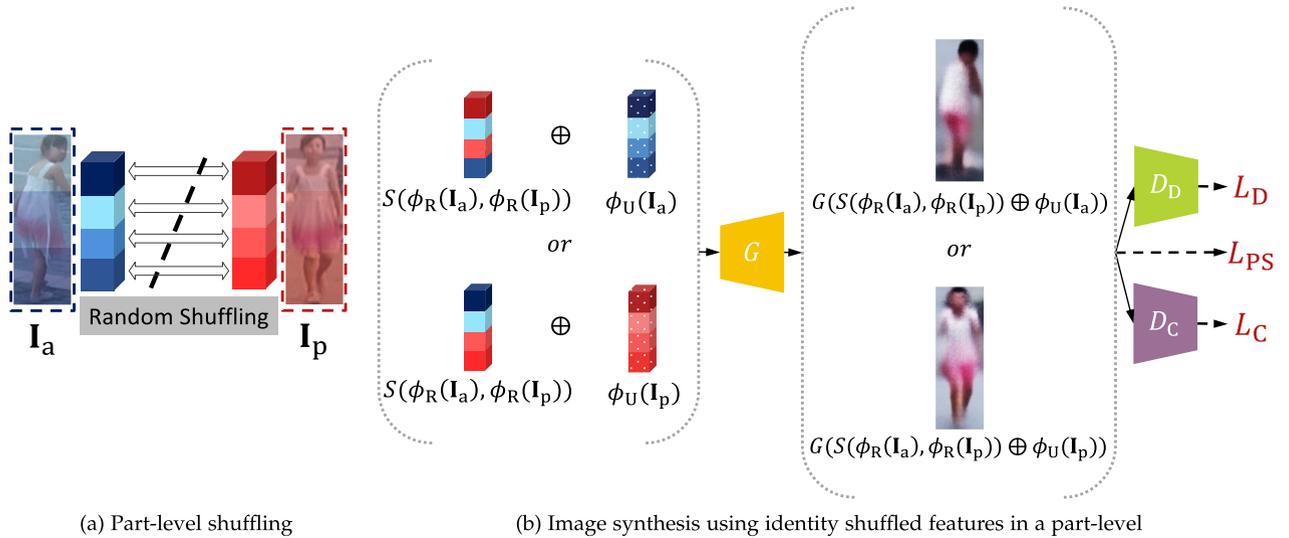

(a) Part-level shuffling        (b) Image synthesis using identity shuffled features in a part-level

Fig. 3. Part-level shuffling. (a) We randomly swap local features between anchor and positive images. (b) Similar to Fig. 2c, we generate person images with identity-related features but shuffled in a part-level and identity-unrelated ones. The part-level shuffling technique makes our model see various combinations of identity-related features, and encourages feature consistencies between corresponding object parts. (Best viewed in color.).

$$\mathcal{L}_{R} = -\sum_{c=1}^{C}\sum_{k=1}^{K} q_c^k \log p(c|\mathbf{w}_c^k \phi_R^k(\mathbf{I}^k)), \tag{1}$$

where $\mathbf{w}_c^k$ is a classifier parameter associated with the identification label $c$ and the region $\mathbf{I}^k$. $q_c^k$ is an index label with $q_c^k = 1$ if the label $c$ corresponds to the identity of the image $\mathbf{I}^k$ (i.e., $c = y$) and $q_c^k = 0$ otherwise. The probability of $\mathbf{I}^k$ with the label $c$ is defined using a softmax function as follows:

$$p(c|\mathbf{w}_c^k \phi_R^k(\mathbf{I}^k)) = \frac{\exp(\mathbf{w}_c^k \phi_R^k(\mathbf{I}^k))}{\sum_{i=1}^{C} \exp(\mathbf{w}_i^k \phi_R^k(\mathbf{I}^k))}. \tag{2}$$

We concatenate all features from three branches, and use it as an identity-related feature for the image $\mathbf{I}$, that is, $\phi_R(\mathbf{I}) = [\phi_R^1(\mathbf{I}^1), \phi_R^2(\mathbf{I}^2), \ldots, \phi_R^K(\mathbf{I}^K)]$.

### 3.3 IS-GAN

The identity-related feature $\phi_R(\mathbf{I})$ from the encoder $E_R$ contains information useful for person reID, such as clothing, length of clothing/hair, and gender. However, the feature $\phi_R(\mathbf{I})$ learned using the classification loss in Eq. (1) only may encode other attributes that are not related to or even distract specifying persons (e.g., human pose, background clutter, scale). This indicates that using the encoder $E_R$ alone is not enough to handle these factors of variations. To address this problem, we use an additional encoder $E_U$ to extract the identity-unrelated feature $\phi_U(\mathbf{I})$, and train the encoders, $E_R$ and $E_U$, such that they give disentangled feature representations for identifying persons. The key idea behind the feature disentanglement is to distill identity-unrelated information from the identity-related feature, and vice versa. To this end, we propose to leverage image synthesis using an identity shuffling technique. Applying this to the whole body and its parts regularizes the disentangled features. Two discriminators, $D_C$ and $D_D$, allow to generate realistic person images of particular identities, further regularizing the disentanglement process.

#### 3.3.1 Identity-Shuffling Loss

We assume that the disentangled person representations satisfy the following conditions: 1) An original image should be reconstructed from its identity-related and -unrelated features; 2) The shared information between different images of the same identity corresponds to the identity-related feature. To implement this, the generator $G$ is required to reconstruct an anchor image $\mathbf{I}_a$ from $\phi_R(\mathbf{I}_a) \oplus \phi_U(\mathbf{I}_a)$ and $\phi_R(\mathbf{I}_p) \oplus \phi_U(\mathbf{I}_a)$, while synthesizing a positive image $\mathbf{I}_p$ from $\phi_R(\mathbf{I}_p) \oplus \phi_U(\mathbf{I}_p)$ and $\phi_R(\mathbf{I}_a) \oplus \phi_U(\mathbf{I}_p)$ (Figs. 2b and 2c). We define an identity-shuffling loss as follows:

$$\mathcal{L}_{S} = \sum_{i,j \in \{a,p\}} \|\mathbf{I}_i - G(\phi_R(\mathbf{I}_j) \oplus \phi_U(\mathbf{I}_i))\|_1. \tag{3}$$

The generator reconstructs the original image when $i = j$, enforcing the combination of identity-related and -unrelated features from the same image to contain all information. When $i \neq j$, it encourages the encoder $E_R$ to extract similar identity-related features, $\phi_R(\mathbf{I}_a)$ and $\phi_R(\mathbf{I}_p)$, from a pair of $\mathbf{I}_a$ and $\mathbf{I}_p$, focusing on the consistent information between them.

#### 3.3.2 Part-Level Shuffling Loss

We also apply the identity shuffling technique to part-level features (Fig. 3). We randomly choose local features from $\phi_R(\mathbf{I}_a)$, and swap them with corresponding ones from $\phi_R(\mathbf{I}_p)$ at the same locations, and vice versa (Fig. 3a). This assumes that horizontal regions in person images contain discriminative body parts sufficient for distinguishing its identity. Similar to Eq. (3), we compute the discrepancies between the original image and its reconstruction using the identity-related features shuffled in a part-level and the identity-unrelated ones (Fig. 3b). Concretely, we define a part-level shuffling loss as follows:

$$\mathcal{L}_{PS} = \sum_{\substack{i,j \in \{a,p\} \\ i \neq j}} \|\mathbf{I}_i - G(S(\phi_R(\mathbf{I}_i), \phi_R(\mathbf{I}_j)) \oplus \phi_U(\mathbf{I}_i))\|_1, \tag{4}$$

where we denote by $S$ a region-wise shuffling operator. The part-level identity shuffling has the following advantages: 1) It enables our model to see various combinations of identity-related features for individual body parts, regularizing a feature disentanglement process; 2) It imposes feature consistencies between corresponding parts of the images.

### 3.3.3 Identity-Unrelated Loss

We disentangle identity-related/-unrelated features without corresponding supervisory signals. Although we train the encoders separately to extract these features, the generator $G$ may largely rely on the identity-unrelated features to synthesize new person images in Eqs. (3) and (4), while ignoring the identity-related ones, which distracts the feature disentanglement process. To circumvent this issue, we use two regularization techniques. In the following, using a slight abuse of notation, we will represent $\phi_R$ and $\phi_U$ random variables of identity-related and -unrelated features, respectively, for notational conciseness.

First, to encourage the distribution of identity-unrelated features to be close to the normal distribution $\mathcal{N}(\mathbf{0}, \mathbf{I})$, and formulate this using a KL divergence loss as follows:

$$\mathcal{L}_U = \sum_{k=1}^{K} D_{KL}(\phi_U^k(\mathbf{I}^k)||\mathcal{N}(\mathbf{0}, \mathbf{I})),\tag{5}$$

where $D_{KL}(p||q) = -\int p(z)log\frac{p(z)}{q(z)}$. The KL divergence loss regularizes the identity-unrelated features by limiting the distribution range, such that they do not contain much identity-related information [24], [25], [26]. This enforces the generator $G$ to use the identity-related features to synthesize new person images, facilitating the disentanglement process.

Second, we instead minimize Pearson correlation coefficients in order to encourage identity-related/-unrelated features to be uncorrelated as follows:

$$\mathcal{L}_U = \sum_{k=1}^{K} \frac{(\phi_R^k(\mathbf{I}^k) - \mu_R^k)(\phi_U^k(\mathbf{I}^k) - \mu_U^k)}{\sigma_R^k \sigma_U^k},\tag{6}$$

where $\mu_R^k$, $\mu_U^k$ and $\sigma_R^k$, $\sigma_U^k$ are the mean and standard deviation of local features, $\phi_R^k$ and $\phi_U^k$, respectively. The decorrelation loss minimizes mutual information between identity-related and -unrelated features [55], and encourages the identity-related and -unrelated encoders to extract different features. Note that it is infeasible to track the mean and standard deviation in Eq. (6) over a dataset during training, where the distributions of the identity-related/-unrelated features vary at every iteration (i.e., a non-stationary process). A naive solution is to compute batch-wise mean and standard deviation of features [56], assuming sampled mini-batches are enough to approximate the distribution of the entire training set. This assumption is reasonable in small and simple datasets such as MNIST [57], but reID benchmarks [28], [29], [30], [48] typically contain a large number of person images with diverse pose variations and distracting scene details. We instead propose to use moving mean and moving standard deviation, which has shown the effectiveness on approximating mean and standard deviation of an entire set in a non-stationary process [58].

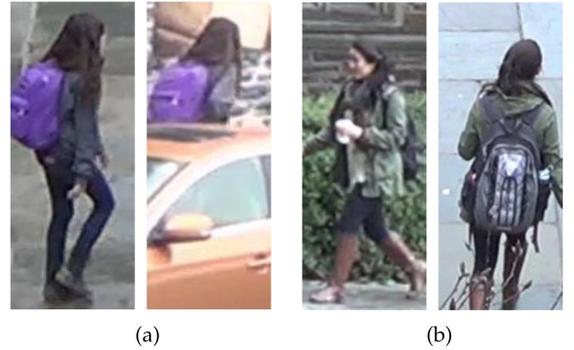

Fig. 4. Examples of image pairs, where shared attributes are not sufficient for determining an identity for reID, due to (a) severe occlusion and (b) viewpoint change.

*Discussion.* We analyze the effect of two alternatives, KL and decorrelation losses, along with the identity-shuffling within our framework. Let us assume that the amount of information for a combination of identity-related and -unrelated features is the same, even identity-related features are swapped between anchor and positive images, as follows:

$$H(\phi_R(\mathbf{I}_a), \phi_U(\mathbf{I}_a)) = H(\phi_R(\mathbf{I}_p), \phi_U(\mathbf{I}_a)),\tag{7}$$

where we denote by $H$ entropy. This can be represented as

$$\begin{aligned}H(\phi_R(\mathbf{I}_a)) - I(\phi_R(\mathbf{I}_a), \phi_U(\mathbf{I}_a)) \\= H(\phi_R(\mathbf{I}_p)) - I(\phi_R(\mathbf{I}_p), \phi_U(\mathbf{I}_a)),\end{aligned}\tag{8}$$

where $I$ is mutual information between features.

The main difference between the KL divergence and decorrelation losses is whether they admit mutual information between identity-related and -unrelated features or not. The decorrelation loss enforces the mutual information between these features, $I(\phi_R(\mathbf{I}_a), \phi_U(\mathbf{I}_a))$ and $I(\phi_R(\mathbf{I}_p), \phi_U(\mathbf{I}_a))$, to be zero explicitly [55], suggesting that it makes the entropy of identity-related features, $\phi_R(\mathbf{I}_a)$ and $\phi_R(\mathbf{I}_p)$, to be identical, i.e., $H(\phi_R(\mathbf{I}_a)) = H(\phi_R(\mathbf{I}_p))$. This helps to extract similar identity-related features from person images of the same identity, which can boost the reID performance. However, if there are lots of training pairs that rarely share identity-related attributes, due to e.g., severe occlusion (Fig. 4a) or viewpoint changes (Fig. 4b), the decorrelation loss rather restricts the amount of information identity-related features can contain. On the contrary, the KL divergence loss does not necessarily make the mutual information, $I(\phi_R(\mathbf{I}_a), \phi_U(\mathbf{I}_a))$ and $I(\phi_R(\mathbf{I}_p), \phi_U(\mathbf{I}_a))$, to be zero. That is, there could be shared information between identity-related and -unrelated features, and thus identity-related features, $\phi_R(\mathbf{I}_a)$ and $\phi_R(\mathbf{I}_p)$, may be different. This prevents our model from removing identity-unrelated information from the identity-related features completely, but gives a better tolerance to the hard cases in Fig. 4.

### 3.3.4 Domain and Class Losses

To train the generator $G$ in Eqs. (3) and (4), we use two discriminators $D_D$ and $D_C$. The domain discriminator $D_D$ [20] helps the generator $G$ to synthesize more realistic person images, and the class discriminator $D_C$ [59] encourages the synthesized images to have the same identification labels of anchor and positive images, further regularizing the feature learning process. Concretely, we define a domain loss $\mathcal{L}_D$ as follows:

$$
\begin{aligned}
\mathcal{L}_D = &\sum_{i \in \{a,p\}} \log D_D(\mathbf{I}_i) \\
&+ \sum_{i,j \in \{a,p\}} \log \left(1 - D_D(G(\phi_R(\mathbf{I}_j) \oplus \phi_U(\mathbf{I}_i)))\right) \\
&+ \sum_{\substack{i,j \in \{a,p\} \\ i \neq j}} \log \left(1 - D_D(G(S(\phi_R(\mathbf{I}_i), \phi_R(\mathbf{I}_j)) \oplus \phi_U(\mathbf{I}_i)))\right).
\end{aligned}
\tag{9}
$$

The domain discriminator $D_D$ is trained, such that it distinguishes real and fake images while the generator $G$ tries to synthesize more realistic images to fool $D_D$. A class loss $\mathcal{L}_C$ is defined as follows:

$$
\begin{aligned}
\mathcal{L}_C = &- \sum_{i \in \{a,p\}} \log D_C(\mathbf{I}_i) \\
&- \sum_{i,j \in \{a,p\}} \log \left(D_C(G(\phi_R(\mathbf{I}_j) \oplus \phi_U(\mathbf{I}_i)))\right) \\
&- \sum_{\substack{i,j \in \{a,p\} \\ i \neq j}} \log \left(D_C(G(S(\phi_R(\mathbf{I}_i), \phi_R(\mathbf{I}_j)) \oplus \phi_U(\mathbf{I}_i)))\right).
\end{aligned}
\tag{10}
$$

The class discriminator $D_C$ classifies identification labels of generated and input person images. When the generator $G$ synthesizes a hard-to-classify image without sufficient identity-related information, the class discriminator $D_C$ would be confused to determine the identification label of the generated image. The generator $G$ thus tries to synthesize a person image of a particular identity associated with the identity-related features, $\phi_R(\mathbf{I}_j)$ and $S(\phi_R(\mathbf{I}_i), \phi_R(\mathbf{I}_j))$.

### 3.3.5  Training Loss

The overall objective is a weighted sum of all loss functions defined as

$$
\begin{aligned}
\min_{E_R,E_U,G,D_C} \max_{D_D} \ &\mathcal{L}(E_R, E_U, G, D_D, D_C) \\
= \ &\lambda_R \mathcal{L}_R + \lambda_U \mathcal{L}_U + \lambda_S \mathcal{L}_S + \lambda_{PS} \mathcal{L}_{PS} + \lambda_D \mathcal{L}_D + \lambda_C \mathcal{L}_C,
\end{aligned}
\tag{11}
$$

where $\lambda_R$, $\lambda_U$, $\lambda_S$, $\lambda_{PS}$, $\lambda_D$, and $\lambda_C$ are weighting factors for each loss. We use either the loss in Eqs. (5) or (6) to train IS-GAN. We denote by IS-GAN$_{KL}$ and IS-GAN$_{DC}$ our models using a KL divergence loss Eq. (5) and a decorrelation loss Eq. (6), respectively.

## 4  EXPERIMENTS

In this section, we evaluate our models, IS-GAN$_{KL}$ and IS-GAN$_{DC}$, for both short-term and long-term reID tasks on standard benchmarks, and provide a detailed analysis including extensive ablation studies.

### 4.1  Implementation Details

#### 4.1.1  Network Architecture

We exploit a ResNet-50 [18] trained for ImageNet classification [19]. Specifically, we use the network cropped at conv4 − 1 as our backbone to extract CNN features. On top of that, we add two heads for identity-related/-unrelated encoders, respectively. Each encoder has part-1, part-2, and part-3 branches that consist of two convolutional, global max pooling, and bottleneck layers but with different numbers of channels and network parameters. The part-1, part-2, and part-3 branches in the encoders give feature maps of size $1 \times 1 \times p$, $1 \times 1 \times 3p$, and $1 \times 1 \times 4p$, respectively. See Section 3.2 for details. We set the size of $p$ (i.e., the number of channels) to 256 for both identity-related and -unrelated encoders. We concatenate all features from three branches for each encoder, and obtain identity-related/-unrelated features. The generator consists of a series of six transposed convolutional layers with batch normalization [60], Leaky ReLU [61] and Dropout [62]. It inputs an addition of identity-related and -unrelated features, a noise vector, and a one-hot encoding of identification labels whose dimensions are 2048, 128, and $C$, respectively. The domain and class discriminators share five blocks consisting of a convolutional layer of stride 2 with instance normalization [63] and Leaky ReLU [61], but have different heads. For the domain discriminator, we add two more blocks, resulting in a features map of size $12 \times 4$. We then use this as an input to PatchGAN [64]. For the class discriminator, we add one more block followed by a fully connected layer.

Note that the IS-GAN$_{KL}$ predicts the distribution of identity-unrelated features (i.e., mean and variance), and samples the features using a reparameterization trick [65], in contrast to IS-GAN$_{DC}$ that directly outputs the features. This requires IS-GAN$_{KL}$ to have additional fully connected layers for estimating the mean and variance of identity-unrelated features, suggesting that it uses extra 83M parameters compared to IS-GAN$_{DC}$. Note also that IS-GAN$_{KL}$ shares the same network architecture with IS-GAN [27], the earlier version of our model, except two minor differences: For IS-GAN$_{KL}$, 1) the dimension of identity-unrelated features is set to 2,048 instead of 512, and 2) a generator inputs the addition of identity-related and -unrelated features rather than the concatenation of them. This enables generating images with identity-related or -unrelated features (e.g., Figs. 9 and 12), which is important to infer what information each feature encodes.

#### 4.1.2  Dataset and Evaluation Metric

We evaluate our models for both short-term and long-term reID tasks. We use Market-1501 [28], CUHK03 [29], and DukeMTMC-reID [30] for the short-term reID, and Celeb-reID [48] for the long-term one. The Market-1501 dataset [28] contains 1,501 pedestrian images captured from six viewpoints. Following the standard split [28], we use 12,936 images of 751 identities for training and 19,732 images of 750 identities for testing. The CUHK03 dataset [29] contains 14,096 images of 1,467 identities captured by two cameras. For the training/testing splits, we follow the experimental protocol in [66]. The DukeMTMC-reID dataset [30], a subset of the DukeMTMC [67], provides 36,411 images of 1,812 identities captured by eight cameras, including 408 identities (distractor IDs) that appear in only one camera. We use the training/test splits provided by [30] corresponding 16,522 images of 702 identities for training and 2,228 query and 17,661 gallery images of 702 identities for testing. The Celeb-reID dataset [48] contains 34,186 images of 1,052 celebrities. Following the standard protocol in [48], we use 20,208 images of 632 identities for training and 13,978 images of 420 identities for testing. We use identity-related features only for evaluation, and measure mean average precision (mAP) and cumulative matching characteristics (CMC) at rank-1.

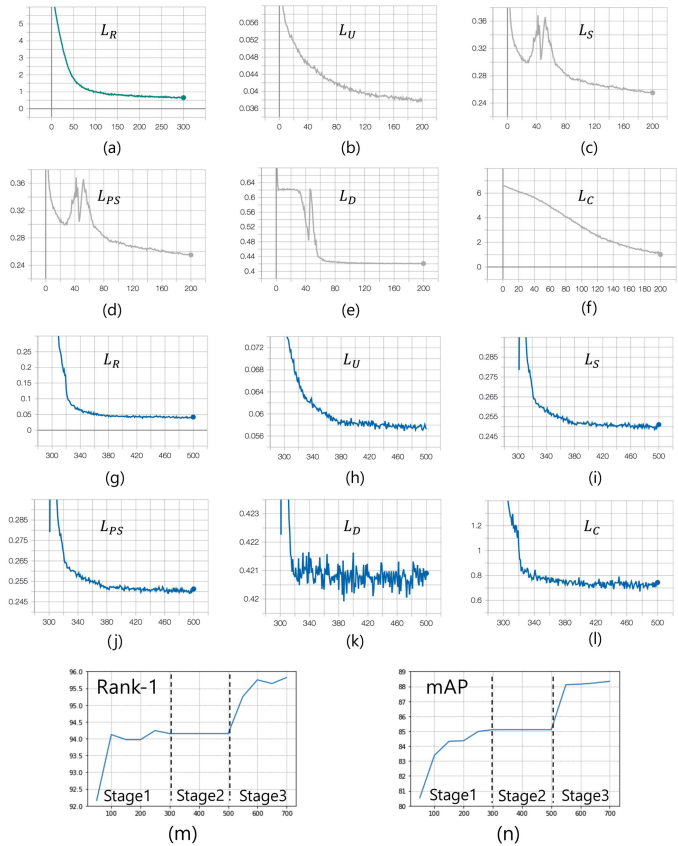

Fig. 5. Visualization of training loss curves on (a) the first, (b-f) the second, and (g-l) the third stages, and (m) rank-1(%) and (n) mAP(%).

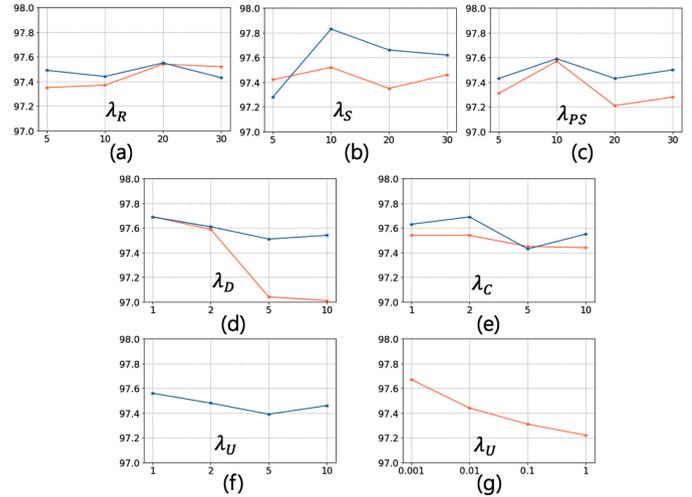

Fig. 6. Sensitivity analysis on hyperparameters of IS-GAN$_{KL}$ (orange) and IS-GAN$_{DC}$ (blue) for (a) identity-related, (b) identity-shuffling, (c) part-level shuffling, (d) domain, (e) class, and (f-g) identity-unrelated losses.

### 4.1.3 Training

To train encoders and a generator, we use the Adam [68] optimizer with $\beta_1 = 0.9$ and $\beta_2 = 0.999$. For discriminators, we use the stochastic gradient descent with a momentum of 0.9. Similar to the training strategy in [40], we train our models in three stages: In the first stage, we train the identity-related encoder $E_R$ using the loss function $\mathcal{L}_R$, which corresponds to the baseline model, for 300 epochs over the training data. A learning rate is set to 2e-4. In the second stage, we fix the baseline, and train the identity-unrelated encoder $E_U$, the generator $G$, and the discriminators $D_D$ and $D_C$ with the corresponding losses $\mathcal{L}_U$, $\mathcal{L}_S$, $\mathcal{L}_{PS}$, $\mathcal{L}_D$, and $\mathcal{L}_C$. This process iterates for 200 epochs with a learning rate of 2e-4. Finally, we train the whole network end-to-end with the learning rate of 2e-5 for 200 epochs. For Celeb-reID [48], motivated by [48], we reduce training epochs for the first and third stages to 50 to prevent an overfitting problem. We plot in Fig. 5 training losses and accuracies of IS-GAN$_{DC}$ for each training stage on Market-1501 [28]. We can observe that the losses slowly decrease over epochs, and rank-1 and mAP reach the highest value when the training phase is finished. Note that rank-1 and mAP remain unchanged during the second stage, since we freeze the parameters of the identity-related encoder. We resize all images into $384 \times 128$, and augment them with random horizontal flipping, cropping, and erasing [69]. For mini-batch, we randomly select 4 different identities, and sample a set of 4 images for each identity. We use label smoothing, cosine annealing, and a warm-up strategy, following [70], [71], [72]. Motivated by the feature fusion strategy in [48], we do not shuffle local features from lower body parts (i.e., the last horizontal slice of part-2 and -3 branches) for Celeb-reID. Note that recognizing identity-related/-unrelated attributes from lower body parts is extremely challenging even for human, especially when clothing and attributes are changed. The training phase takes 8 hours in total with an RTX2080Ti GPU, and requires 10 GB memory.

### 4.1.4 Hyperparameter

For a KL divergence loss, we empirically find that initial training with a large value of $\lambda_U$ is unstable. We thus set $\lambda_U$ to 1e-3 in the second stage, and increase it to 1e-2 in the third stage to regularize the feature disentanglement. For a decorrelation loss, we set $\lambda_U$ to 1. Following [25], [40], we fix $\lambda_S$ and $\lambda_D$ to 10 and 1, respectively. To set other parameters, we randomly split IDs in the training dataset of Market-1501 [28] into 651/100, and use corresponding images as training/validation samples. We then sample 160 query images from the validation split, setting the remaining images to a gallery set. We use a grid search to set the parameters ($\lambda_R = 20$, $\lambda_{PS} = 10$, $\lambda_C = 2$) with $\lambda_R \in \{5, 10, 20\}$, $\lambda_{PS} \in \{5, 10, 20\}$, and $\lambda_C \in \{1, 2\}$ on the validation split. We fix all parameters, and train our models with the same parameters on Market-1501 [28], CUHK03 [29], DukeMTMC-reID [30], and Celeb-reID [48]. We show in Fig. 6 rank-1 accuracies(%) of IS-GAN$_{KL}$ and IS-GAN$_{DC}$ on the validation split of Market-1501 [28] while varying hyperparameters. The default hyperparameters are chosen with the aforementioned setting as follows: $\lambda_R = 20$, $\lambda_S = 10$, $\lambda_{PS} = 10$, $\lambda_D = 1$, $\lambda_C = 2$, $\lambda_U = 1$ (IS-GAN$_{DC}$),

TABLE 1
Quantitative Comparison With the State of the Art on Market-1501 [28], CUHK03 [29], DukeMTMC-reID [30], and Celeb-reID [48] in Terms of Rank-1 Accuracy(%) and mAP(%)

| Methods | f-dim | Market-1501 | | CUHK03 | | | | DukeMTMC-reID | | Celeb-reID | |
|---|---|---|---|---|---|---|---|---|---|---|---|
| | | | | labeled | | detected | | | | | |
| | | R-1 | mAP | R-1 | mAP | R-1 | mAP | R-1 | mAP | R-1 | mAP |
| IDE [46] | 2,048 | 73.9 | 47.8 | 22.2 | 21.0 | 21.3 | 19.7 | - | - | 42.9 | 5.9 |
| SVDNet [73] | 2,048 | 82.3 | 62.1 | 40.9 | 37.8 | 41.5 | 37.3 | 76.7 | 56.8 | - | - |
| DaRe [74] | 128 | 88.5 | 74.2 | 64.5 | 60.2 | 61.6 | 58.1 | 79.1 | 63.0 | - | - |
| ReIDCaps [48] | 1,024 | 89.0 | 72.7 | - | - | - | - | 81.2 | 62.6 | <u>51.2</u> | 9.8 |
| PN-GAN [21] | 1,024 | 89.4 | 72.6 | - | - | - | - | 73.6 | 53.2 | - | - |
| MLFN [47] | 1,024 | 90.0 | 74.3 | 54.7 | 49.2 | 52.8 | 47.8 | 81.0 | 62.8 | 41.4 | 6.0 |
| FD-GAN [40] | 2,048 | 90.5 | 77.7 | - | - | - | - | 80.0 | 64.5 | - | - |
| HA-CNN [15] | 1,024 | 91.2 | 75.7 | 44.4 | 41.0 | 41.7 | 38.6 | 80.5 | 63.8 | 47.6 | 9.5 |
| PGFA [34] | 12,288 | 91.2 | 76.8 | - | - | - | - | 82.6 | 65.5 | - | - |
| Part-Aligned [32] | 512 | 91.7 | 79.6 | - | - | - | - | 84.4 | 69.3 | 19.4 | 6.4 |
| PCB [11] | 12,288 | 92.3 | 77.4 | - | - | 61.3 | 54.2 | 81.8 | 66.1 | - | - |
| PyrNet [75] | 512 | 93.6 | 81.7 | 71.6 | 68.3 | 68.0 | 63.8 | 87.1 | 74.0 | - | - |
| PCB+RPP [11] | 12,288 | 93.8 | 81.6 | - | - | 63.7 | 57.5 | 83.3 | 69.2 | 37.1 | 8.2 |
| AANet [76] | 2,048 | 93.9 | 83.4 | - | - | - | - | 87.7 | 74.3 | - | - |
| HPM [77] | 3,840 | 94.2 | 82.7 | - | - | 63.9 | 57.5 | 86.6 | 74.3 | - | - |
| CASN+PCB [78] | 12,288 | 94.4 | 82.8 | 73.7 | 68.0 | 71.5 | 64.4 | 87.7 | 73.7 | - | - |
| BoT [70] | 2,048 | 94.5 | 85.9 | - | - | - | - | 86.4 | 76.4 | - | - |
| DG-Net [31] | 1,024 | 94.8 | 86.0 | - | - | - | - | 86.6 | 74.8 | - | - |
| OS-Net [45] | 512 | 94.8 | 84.9 | - | - | 72.3 | 67.8 | 88.6 | 73.5 | - | - |
| Top-DB-Net [71] | 2,048 | 94.9 | 85.8 | 79.4 | 75.4 | 77.3 | 73.2 | 87.5 | 73.5 | - | - |
| MHN-6 [35] | 12,288 | 95.1 | 85.0 | 77.2 | 72.4 | 71.7 | 65.4 | 89.1 | 77.2 | - | - |
| P²-Net [33] | 256 | 95.2 | 85.6 | 78.3 | 73.6 | 74.9 | 68.9 | 86.5 | 73.1 | - | - |
| AdaptiveL2 [72] | 2,048 | 95.3 | 88.3 | - | - | - | - | 88.9 | 79.9 | - | - |
| ABD-Net [36] | 2,048 | 95.6 | 88.3 | - | - | - | - | 89.0 | 78.6 | - | - |
| DNDM [79] | 2,048 | 95.6 | 87.1 | - | - | - | - | 88.8 | 78.7 | - | - |
| MGN [12] | 2,048 | 95.7 | 86.9 | 68.0 | 67.4 | 66.8 | 66.0 | 88.7 | 78.4 | 49.0 | 10.8 |
| Pyramid [80] | 2,048 | 95.7 | 88.2 | 78.9 | <u>76.9</u> | **78.9** | **74.8** | 89.0 | 79.0 | - | - |
| SCAL [37] | 2,048 | <u>95.8</u> | <u>89.3</u> | 74.8 | 72.3 | 71.1 | 68.6 | 88.9 | 79.1 | - | - |
| Circleloss [81] | 2,048 | **96.1** | 87.4 | - | - | - | - | - | - | - | - |
| IS-GAN$_{KL}$ | 2,048 | 95.7 | 88.1 | <u>79.5</u> | **77.3** | 76.7 | <u>73.9</u> | **91.4** | **80.9** | **54.5** | **12.8** |
| IS-GAN$_{DC}$ | 2,048 | **96.1** | **89.4** | **80.0** | **77.3** | <u>77.4</u> | <u>73.9</u> | <u>90.8</u> | <u>80.3</u> | **54.5** | <u>12.5</u> |
| DaRe [74] + RR | 128 | 90.8 | 85.9 | 72.9 | 73.7 | 69.8 | 71.2 | 84.4 | 79.6 | - | - |
| Part-Aligned [32] + RR | 512 | 93.4 | 89.9 | - | - | - | - | 88.3 | 83.9 | - | - |
| PyrNet [75] + RR | 512 | 94.6 | 91.4 | 80.8 | 82.7 | 77.1 | 78.7 | 90.3 | 87.7 | - | - |
| BoT [70] + RR | 2,048 | 95.4 | 94.2 | - | - | - | - | 90.3 | <u>89.1</u> | - | - |
| Top-DB-Net [71] + RR | 2,048 | 95.5 | 94.1 | <u>86.7</u> | **88.5** | **85.7** | **86.9** | <u>90.9</u> | 88.6 | - | - |
| IS-GAN$_{KL}$ + RR | 2,048 | <u>96.3</u> | <u>94.8</u> | 86.4 | 88.0 | 84.4 | 85.8 | **93.2** | **90.5** | 54.9 | <u>14.5</u> |
| IS-GAN$_{DC}$ + RR | 2,048 | **96.4** | **95.2** | **87.1** | <u>88.4</u> | <u>84.7</u> | <u>86.3</u> | **93.2** | 90.5 | <u>54.0</u> | 14.9 |

*The results of other methods on Market-1501 [28], CUHK03 [29], and DukeMTMC-reID [30] are taken from each paper. The results for Celeb-reID [48] are taken from [48]. Numbers in bold indicate the best performance and underscored ones are the second best. RR: Re-ranking using [66].*

$\lambda_U = 0.001$ (IS-GAN$_{KL}$). The results suggest that our models are fairly robust against the hyperparameters in that IS-GAN$_{KL}$ and IS-GAN$_{DC}$ show rank-1 accuracies higher than 97%, regardless of the change of the parameters.

## 4.2 Results

We show quantitative and qualitative results on short-term and long-term reID benchmarks. We use a single query, and do not use any post-processing techniques (e.g., a re-ranking method [66]).

### 4.2.1 Quantitative Comparison

*Short-Term reID.* We show in Table 1 rank-1 accuracy and mAP for Market-1501 [28], CUHK03 [29] and DukeMTMC-reID [30], and compare our models with the state of the art, including FD-GAN [40], DG-Net [31], MGN [12], P²-Net [33], and SCAL [37]. We can see that IS-GAN$_{DC}$ sets a new state of the art on Market-1501 and CUHK03, achieving

96.1% rank-1 accuracy and 89.4% mAP on Market-1501, and 80.0%/77.4% rank-1 accuracy and 77.3%/73.9% mAP with labeled/detected images on CUHK03. IS-GAN$_{KL}$ shows state-of-the-art performance on DukeMTMC-reID, providing 91.4% rank-1 accuracy and 80.9% mAP. Although IS-GAN$_{DC}$ uses fewer parameters than IS-GAN$_{KL}$, it shows meaningful performance gains compared to IS-GAN$_{KL}$ in terms of rank-1/mAP, i.e., 0.6/0.6, 2.5/3.0, and 2.9/2.1 on Market-1501, CUHK03-detected, and CUHK03-labeled, and shows comparable performance on DukeMTMC-reID.

FD-GAN [40] is similar to ours in that both use a GAN-based distillation technique for person reID. It extracts identity-related and pose-unrelated features using extra pose labels. Distilling other factors except for human pose is, however, infeasible. Our models on the other hand disentangle identity-related/-unrelated features through identity shuffling, factorizing identity-related factors and others irrelevant to person reID, such as pose, scale, background clutter, without corresponding supervisory signals. Accordingly, the identity-related feature is much more robust to such factors of variations than the identity-related and pose-unrelated one of FD-GAN, showing a better performance on Market-1501 and DukeMTMC-reID. Note that the results of FD-GAN on CUHK03 are excluded, as it uses different training/test splits.

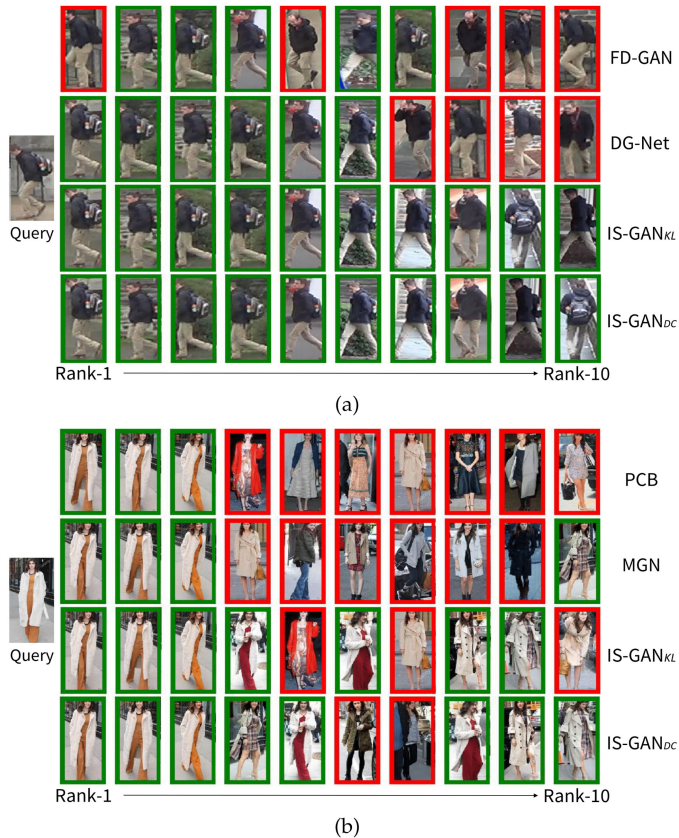

Fig. 7. Visual comparison of retrieval results on (top) DukeMTMC-reID [30] and (bottom) Celeb-reID [48]. We compute the euclidean distances between identity-related features of query and gallery images, and visualize top-10 results sorted according to the distance. The results with green boxes have the same identity as the query, while those with red boxes do not. (Best viewed in color.).

MGN [12] uses a backbone network the same as ours to extract initial part-level features. As it is trained with a hard-triplet loss, the part-level features of MGN capture discriminative attributes of person images well. For Market-1501, MGN shows the reID performance comparable with our models. Note that, compared to other datasets, person images in Market-1501 show fewer pose and attribute variations. The reID performance of MGN, however, drops significantly on other datasets, especially for CUHK03, where the same person is captured with different poses and illumination, demonstrating that the person representations for MGN are not robust to such factors.

P²-Net [33] and SCAL [37] use a part-aligned module and an attention method, respectively. Specifically, P²-Net aggregates pixel-wise features for each human body part, and SCAL guides the network to focus on salient regions of images (e.g., body part or person belongings). They are, however, limited in that person representations are robust to a particular factor (e.g., pose changes or background clutter), and still vulnerable to other factors, such as illumination, occlusion, and scale changes. As a result, while each shows comparable performance with our models on a few benchmarks, there are large performance gaps on others. Note that P²-Net additionally uses a human parsing dataset [82] to train the part-aligned module.

*Long-Term reID*. We also compare in Table 1 our models with the state of the art on Celeb-reID [48]. ReIDCaps [48] exploits a capsule network [49], representing different identities by the length of each capsule, with its orientation being in charge of appearance changes of the same identity. By modeling the visual changes of each identity, this approach gives better results than other methods designed for the short-term reID task. The number of appearance variations is, however, limited by the dimension of capsules. On the contrary, our models are flexible to the number of the variations by factorizing images into identity-related/-unrelated features, and they outperform ReIDCaps by significant margins. Note that, while ReIDCaps degrades the reID performance severely on short-term reID benchmarks, including Market-1501 [28] and DukeMTMC-reID [30], our models outperform the state of the art consistently on standard benchmarks for both short-term and long term reID tasks. Note that our model achieves state of the art on both short- and long-term reID in a unified manner.

### 4.2.2 Qualitative Comparison

*Short-Term reID*. We show in Fig. 7a person retrieval results of FD-GAN [40], DG-Net [31], and ours on DukeMTMC-reID [30]. We can see that FD-GAN and DG-Net mainly focus on color information. For example, they retrieve many person images of different identities but with the outfit of similar color to the query person, such as a black jacket, beige trousers, and gray bag. Note that all retrieved images by DG-Net have a similar pose to the query, indicating it fails to get rid of pose-related information from person representations. In contrast, our models retrieve person images of the same identity as the query correctly, and they are robust to large pose variations, occlusion, background clutter, and scale changes.

*Long-Term reID*. In Fig. 7b, we show reID results of PCB [11], MGN [12], and ours on Celeb-reID [48]. All models retrieve person images with the same clothing as the query in high rank (i.e., rank 1-3) successfully. PCB and MGN, however, fail to re-identify the query person for the case of outfit changes, retrieving person images with different identities but having a similar pose. We can see that our models better handle the long-term reID problem. For example, they retrieve correct images, w.r.t the query person, even with different clothes and belongings.

### 4.2.3 Ablation Study

We show in Table 2 an ablation analysis on different losses in our models. We report rank-1 accuracy and mAP on Market-1501 [28], CUHK03 [29], DukeMTMC-reID [30], and Celeb-reID [48]. We can see that 1) disentangling identity-related/-unrelated features using an identity shuffling technique gives better results on all datasets, but the performance gain for Celeb-reID [48], which contains person images of large appearance variations, is more significant, 2) applying the identity shuffling technique in a part-level further boosts the reID performance, and 3) domain and class discriminators

TABLE 2
Ablation Studies of IS-GAN$_{KL}$ and IS-GAN$_{DC}$ on Market-1501 [28], CUHK03 [29], DukeMTMC-reID [30], and Celeb-reID [48] in Terms of Rank-1 Accuracy(%) and mAP(%)

| | $\mathcal{L}_R$ | $\mathcal{L}_U$ | $\mathcal{L}_S$ | $\mathcal{L}_{PS}$ | $\mathcal{L}_D$ | $\mathcal{L}_C$ | Market-1501 R-1 | mAP | CUHK03-labeled R-1 | mAP | DukeMTMC-reID R-1 | mAP | Celeb-reID R-1 | mAP |
|---|---|---|---|---|---|---|---|---|---|---|---|---|---|---|
| Baseline | ✓ | | | | | | 94.15 | 85.10 | 72.43 | 70.53 | 86.98 | 76.38 | 48.59 | 9.46 |
| IS-GAN$_{KL}$ | ✓ | ✓ | ✓ | | | | 94.80 | 87.04 | 73.71 | 72.51 | 88.33 | 78.88 | 53.03 | 12.53 |
| | ✓ | ✓ | ✓ | ✓ | | | 94.95 | 87.44 | 74.00 | 72.77 | 88.82 | 79.06 | 53.26 | 12.63 |
| | ✓ | ✓ | ✓ | ✓ | ✓ | | 95.10 | 87.52 | 74.21 | 72.95 | 88.73 | 79.10 | 53.43 | 12.68 |
| | ✓ | ✓ | ✓ | ✓ | | ✓ | 95.10 | 87.54 | 72.86 | 72.67 | 88.51 | 79.10 | 53.53 | 12.65 |
| | ✓ | ✓ | ✓ | ✓ | ✓ | ✓ | 95.16 | 87.72 | 74.93 | 72.83 | 89.27 | 79.19 | 53.60 | 12.69 |
| IS-GAN$_{DC}$ | ✓ | ✓ | ✓ | | | | 95.52 | 87.89 | 76.71 | 75.38 | 88.82 | 78.93 | 53.26 | 12.25 |
| | ✓ | ✓ | ✓ | ✓ | | | 95.69 | 88.21 | 77.79 | 75.37 | 89.05 | 78.87 | 53.30 | 12.36 |
| | ✓ | ✓ | ✓ | ✓ | ✓ | | 95.67 | 88.27 | 76.64 | 75.27 | 88.96 | 79.04 | 53.43 | 12.37 |
| | ✓ | ✓ | ✓ | ✓ | | ✓ | 95.64 | 88.11 | 76.71 | 75.19 | 89.09 | 79.03 | 53.10 | 12.40 |
| | ✓ | ✓ | ✓ | ✓ | ✓ | ✓ | 95.75 | 88.30 | 77.43 | 75.76 | 89.14 | 79.08 | 53.43 | 12.50 |

*Numbers in bold indicate the best performance and underscored ones are the second best. $\mathcal{L}_R$: an identity-related loss; $\mathcal{L}_U$: an identity-unrelated loss; $\mathcal{L}_S$: an identity-shuffling loss; $\mathcal{L}_{PS}$: a part-level shuffling loss; $\mathcal{L}_D$: a domain loss; $\mathcal{L}_C$: a class loss.*

TABLE 3
Quantitative Results of IS-GAN$_{DC}$, IS-GAN$_{KL}$, and Their Variants on Market-1501 [28], CUHK03 [29], DukeMTMC-reID [30], and Celeb-reID [48] in Terms of Rank-1 Accuracy(%) and mAP(%)

| | $\mathcal{L}_U$ | mov. | Market-1501 R-1 | mAP | CUHK03-labeled R-1 | mAP | DukeMTMC-reID R-1 | mAP | Celeb-reID R-1 | mAP |
|---|---|---|---|---|---|---|---|---|---|---|
| IS-GAN$_{DC}$ | X | X | 95.5 | 87.8 | 74.3 | 73.2 | 88.6 | 79.1 | 53.1 | 12.1 |
| | ✓ | X | 94.7 | 83.0 | 63.0 | 61.3 | 87.4 | 74.7 | 50.2 | 10.3 |
| | ✓ | ✓ | 95.8 | 88.3 | 77.4 | 75.8 | 89.1 | 79.1 | 53.4 | 12.5 |
| IS-GAN$_{KL}$ | X | X | 95.2 | 87.7 | 71.7 | 70.1 | 89.3 | 79.2 | 53.6 | 12.7 |
| | ✓ | ✓ | 95.3 | 87.8 | 71.4 | 70.4 | 88.7 | 79.2 | 53.7 | 12.6 |

*Numbers in bold indicate the best performance and underscored ones are the second best.*

are complementary, and combining all losses gives the best results. Note that IS-GAN$_{DC}$, which encourages disentangled features to be uncorrelated, performs better than IS-GAN$_{KL}$ on Market-1501 [28] and CUHK03 [29], which contains person images of comparatively less distracting scene details. On the contrary, IS-GAN$_{KL}$ gives better results on challenging datasets, such as DukeMTMC-reID [30] and Celeb-reID [48], where factorizing images into uncorrelated features is ambiguous due to e.g., severe occlusion and/or frequent clothing variations.

### 4.2.4 Losses

*Decorrelation Loss.* We show the effect of a decorrelation loss for IS-GAN$_{DC}$ in Table 3. We compare IS-GAN$_{DC}$ with two variants: One is trained without the decorrelation loss (IS-GAN$_{DC}(w/o\mathcal{L}_U)$), and the other is trained with the decorrelation loss but using batch-wise means and standard deviations (IS-GAN$_{DC}(w/batch-wise\mathcal{L}_U)$). All other settings including the network architecture are the same as IS-GAN$_{DC}$. From the first and third rows, we can see that the decorrelation loss is effective to remove identity-unrelated information from the identity-related features, boosting the retrieval performance for both short-term and long-term reID tasks. We can also see from the second and third rows that the loss rather degrades the reID performance significantly with batch-wise means and standard deviations. This indicates that exploiting moving means and standard deviations is critical to decorrelate identity-related/-unrelated features, where the distributions of the features keep varying over every iteration at training time.

We show in Fig. 9 images generated using IS-GAN$_{DC}$ and the variants. As expected from the results in Table 3, IS-GAN$_{DC}(w/o\mathcal{L}_U)$ and IS-GAN$_{DC}$ show similar results. IS-GAN$_{DC}(w/o\mathcal{L}_U)$, on the other hand, does not regularize features to be uncorrelated. As a result, its identity-related/-unrelated features often share common information, similar to IS-GAN$_{KL}$, such as the background in the blue boxes of Fig. 9. We can see that IS-GAN$_{DC}(w/batch-wise\mathcal{L}_U)$ reconstructs input images even with identity-unrelated features only, indicating that computing the decorrelation loss in a batch-wise manner prevents regularizing a disentanglement process. We can also observe that IS-GAN$_{DC}(w/batch-wise\mathcal{L}_U)$ generates similar images, even using identity-related features from different persons (see the yellow boxes of Fig. 9). This suggests that identity-related features for IS-GAN$_{DC}(w/batch-wise\mathcal{L}_U)$ are not discriminative for person reID, degrading the retrieval performance severely.

We also provide the result when regularizing IS-GAN$_{KL}$ using a decorrelation loss, together with a KL divergence term. Note that we could not regularize IS-GAN$_{DC}$ with a KL divergence loss, since applying the KL divergence loss to identity-unrelated features requires a specialized network

TABLE 4
Quantitative Results of Our Models Using PCB [11]
as a Backbone Network

|  | f-dim | $\mathcal{L}_{PS}$ | R-1 | mAP |
|---|---|---|---|---|
| PCB | 1,536 | ✗ | 91.5 | 77.6 |
| PCB + IS-GAN$_{KL}$ | 1,536 | ✗ | 93.1 | 81.0 |
|  |  | ✓ | 93.3 | 81.7 |
| PCB + IS-GAN$_{DC}$ | 1,536 | ✗ | 93.4 | 81.2 |
|  |  | ✓ | 93.9 | 81.4 |

*We report rank-1 accuracy(%) and mAP(%) on Market-1501 [28].*

architecture for estimating the mean and variance of the features. Identity-unrelated features of IS-GAN$_{KL}$ are randomly sampled from the estimated mean and variance, suggesting that an infinite number of distinct identity-unrelated features could be extracted from a single image. Decorrelating identity-related factors w.r.t all randomly sampled identity-unrelated features thus might not be feasible. Indeed, as shown in Table 3, we do not obtain meaningful gains on reID benchmarks in terms of rank-1 accuracy(%) and mAP(%) by regularizing IS-GAN$_{KL}$ with the decorrelation term.

*Part-Level Shuffling Loss.* To evaluate the generalization ability of our models, we use PCB [11] as our baseline, and add our modules for IS-GAN$_{KL}$ and IS-GAN$_{DC}$ on top of that. We modify the network architecture such that each part-level feature has the size of $1 \times 1 \times 256$ for efficient computation. Note that the original PCB [11] also gives six part-level features, but with the size of $1 \times 1 \times 2048$. We can see from Table 4 that our models improve the performance of PCB [11] consistently. Table 5 shows the effect of a part-level shuffling loss for different numbers of body parts. We can see that 1) the part-level shuffling loss generalizes well across different numbers of body parts, and 2) both models perform better, as more body parts are used.

TABLE 5
Quantitative Results for Different Numbers of Body Parts on
Market-1501 [28] in Terms of Rank-1 Accuracy(%) and mAP(%)

|  |  | f-dim | $\mathcal{L}_{PS}$ | R-1 | mAP |
|---|---|---|---|---|---|
| IS-GAN$_{KL}$ | part-2 | 768 | ✗ | 94.4 | 85.4 |
|  |  |  | ✓ | 94.6 | 85.5 |
|  | part-3 | 1,024 | ✗ | 94.5 | 84.5 |
|  |  |  | ✓ | 94.8 | 84.9 |
|  | part-1,2 | 1,024 | ✗ | 94.5 | 86.9 |
|  |  |  | ✓ | 95.1 | 87.1 |
|  | part-1,3 | 1,280 | ✗ | 94.7 | 86.3 |
|  |  |  | ✓ | 95.1 | 86.9 |
|  | part-1,2,3 | 2,048 | ✗ | 94.9 | 87.4 |
|  |  |  | ✓ | 95.2 | 87.7 |
| IS-GAN$_{DC}$ | part-2 | 768 | ✗ | 94.5 | 85.7 |
|  |  |  | ✓ | 95.0 | 86.3 |
|  | part-3 | 1,024 | ✗ | 94.4 | 85.2 |
|  |  |  | ✓ | 95.7 | 85.5 |
|  | part-1,2 | 1,024 | ✗ | 95.2 | 87.0 |
|  |  |  | ✓ | 95.7 | 87.6 |
|  | part-1,3 | 1,280 | ✗ | 95.2 | 87.0 |
|  |  |  | ✓ | 95.7 | 87.1 |
|  | part-1,2,3 | 2,048 | ✗ | 95.5 | 88.1 |
|  |  |  | ✓ | 95.8 | 88.3 |

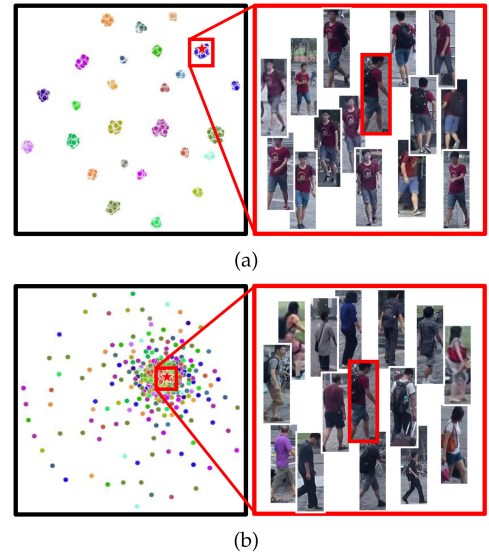

(a)

(b)

Fig. 8. t-SNE [83] visualization for (a) identity-related features and (b) identity-unrelated features on Market-1501 [28]. We randomly sample 26 identities from the test split. The points with the same color indicate the same identity. (Best viewed in color.).

## 4.3 Discussion

### 4.3.1 Visual Analysis on Short-Term reID

*Identity Shuffling.* Fig. 8 shows the t-SNE plots for identity-related and -unrelated features of IS-GAN$_{KL}$ on the test split of Market-1501 [28], where we randomly sample 26 identities. IS-GAN$_{DC}$ shows similar results, so we omit them. We can clearly see that the identity-related features for person images with the same identities are closely embedded, forming compact clusters. On the contrary, the identity-unrelated features are scattered, while the corresponding images with similar poses and background clutter (e.g., concrete pavers) are embedded into close regions.

We visualize in Fig. 9 reconstructed images by IS-GAN$_{KL}$ (Fig. 9a) and IS-GAN$_{DC}$ (Fig. 9b), when a generator $G$ inputs the addition of identity-related/-unrelated features $\phi_R(\mathbf{I}) \oplus \phi_U(\mathbf{I})$, identity-related features $\phi_R(\mathbf{I})$, and identity-unrelated features $\phi_U(\mathbf{I})$, respectively. We can observe that when our models input identity-related features alone, they generate pose-normalized images that have the same clothing color as the original ones, indicating pose information is successfully removed from the identity-related features. In addition, the green boxes in Fig. 9 show that identity-related features are independent of scale changes. On the other hand, when our models input identity-unrelated features alone, they reconstruct decolorized images with the same pose, scale, and background as the original ones, suggesting that these factors are encoded in identity-unrelated features. Note that identity-related/-unrelated features for IS-GAN$_{KL}$ sometimes share common information, since they would not necessarily be uncorrelated. (See the background in the red boxes of Fig. 9 for the example.)

*Part-Level Shuffling.* Fig. 10 visualizes the ability of IS-GAN$_{KL}$ to disentangle identity-related/-unrelated features in a part-level. We omit results of IS-GAN$_{DC}$, as they show similar results. We shuffle the identity-related/-unrelated features for upper/lower parts between person images of different identities. When identity-related features are shuffled e.g., in the upper left picture, we can see that our model changes colors of T-shirts between persons but with the same pose and background. This suggests that the identity-related features do not contain pose and background information. Interestingly, when identity-unrelated features are shuffled, our model generates new images, where background and pose information for the corresponding parts are changed. For example, in the upper right picture, the person looking at the front side now sees the left side and vice versa, when shuffling the features between upper parts, while preserving the shapes of the legs in the lower parts.

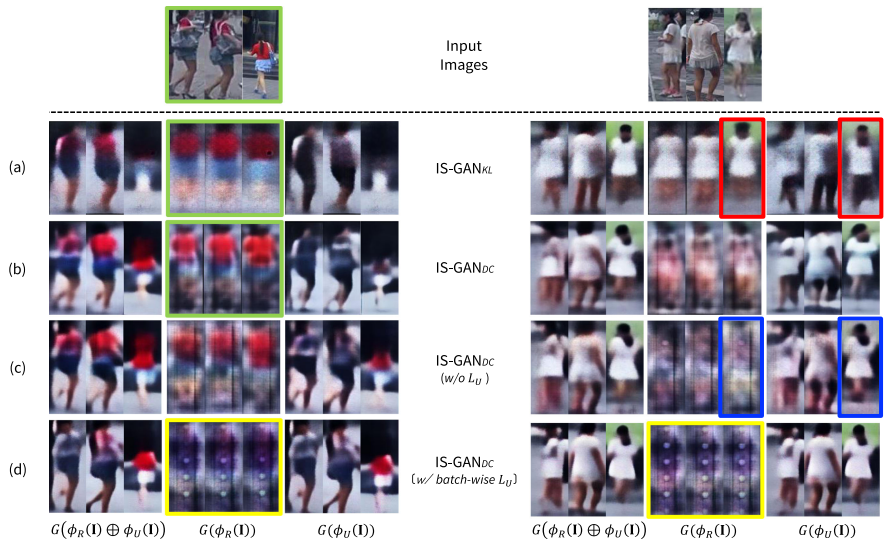

Fig. 9. Visual comparison of reconstructed images using feature representations obtained from (a) IS-GAN$_{KL}$, (b) IS-GAN$_{DC}$, (c) IS-GAN$_{DC}(w/o\mathcal{L}_U)$, and (d) IS-GAN$_{DC}(w/batch-wise\mathcal{L}_U)$. For each model, a generator inputs an addition of identity-related and -unrelated features, identity-related ones, or identity-unrelated ones, respectively. (Best viewed in color.).

### 4.3.2 Visual Analysis on Long-Term reID

*Identity Shuffling.* We visualize in Fig. 11 t-SNE embeddings of identity-related (Fig. 11a) and -unrelated (Fig. 11b) features of IS-GAN$_{KL}$ trained on Celeb-reID. We randomly sample person images of 46 identities. We omit results of IS-GAN$_{DC}$, as they show similar results. We can see that although a certain person (Clark Gregg, a hollywood actor) changes his outfits, identity-related features extracted from his images are still closely embedded. For identity-unrelated features, on the other hand, person images of celebrities, who wear similar color clothing (e.g., a black jacket with black pants) but have different identities, are embedded into near regions. This demonstrates that identity-unrelated features encode clothing colors for the long-term reID task, which is contrary to the short-term reID task (Fig. 8). We visualize in Fig. 11c t-SNE embeddings of identity-related and -unrelated features, where the color of the points indicates gender. Interestingly, the embeddings of identity-related features are almost linearly separated w.r.t gender, while identity-unrelated ones are not. This further demonstrates that identity-related features encode gender information.

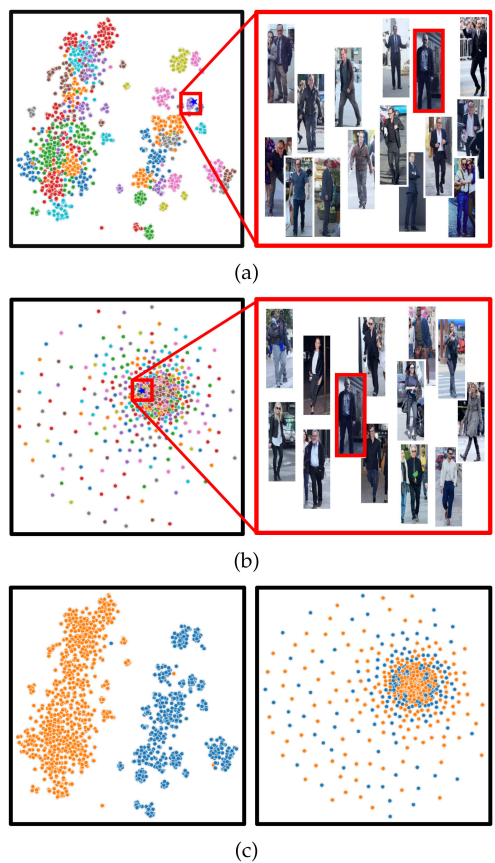

Fig. 11. t-SNE visualizations for (a) identity-related features and (b) identity-unrelated ones on Celeb-reID. We randomly sample 46 identities from the test split. The points with the same color indicate the same identity. We manually label gender annotations for the sampled identities and show (c) t-SNE embeddings of identity-related and -unrelated features. Here, the color represents gender labels (blue: male, orange: female). (Best viewed in color.).

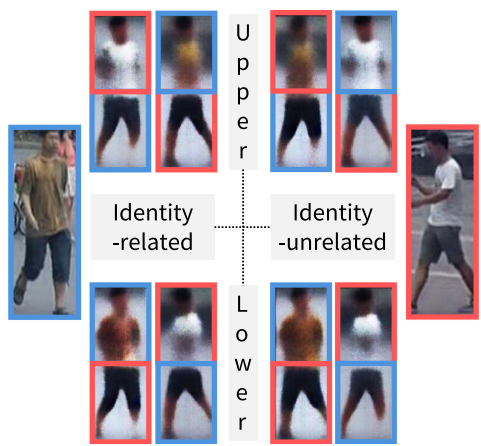

Fig. 10. An example of generated images using a part-level identity shuffling technique. We generate person images by shuffling identity-related/-unrelated features for lower/upper body parts between two images. We shuffle either identity-related or -unrelated features while fixing the other. (Best viewed in color.).

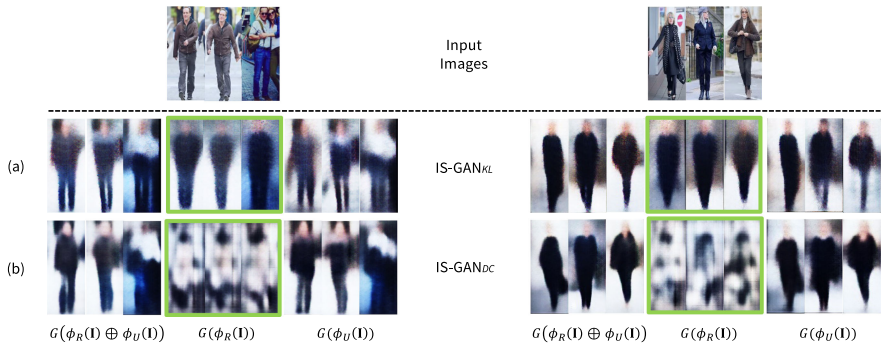

Fig. 12. Visual comparison of reconstructed images using feature representations obtained from (a) IS-GAN$_{KL}$ and (b) IS-GAN$_{DC}$ on Celeb-reID. For each model, a generator inputs an addition of identity-related and -unrelated features, identity-related ones, or identity-unrelated ones, respectively. (Best viewed in color.)

Fig. 12 shows reconstructed images by IS-GAN$_{KL}$ (Fig. 12a) and IS-GAN$_{DC}$ (Fig. 12b), where the generator inputs the addition of identity-related/-unrelated features, identity-related features, and identity-unrelated features, respectively. We can observe that similar images with a normalized pose are reconstructed regardless of person outfits, when generators take identity-related features only, as in the green boxes. The output images generated using identity-unrelated features, on the other hand, show the same background and clothing color as those using the addition of identity-related/-unrelated features. Note that person images of the same identity do not share most information other than, e.g., gender or face color, in a long-term reID task. The reconstructed images from identity-unrelated features are thus fairly similar to those from the addition of identity-related/-unrelated features.

*Part-Level Shuffling.* We show in Fig. 13 the generated images when a part-level shuffling technique is applied to upper body parts. Since we do not shuffle local features from lower body parts to train our model on a long-term reID dataset, we omit the corresponding results. Please refer to Paragraph 4.1.3 for details. We can see that shuffling identity-related features from the upper body parts of two images changes face color between persons. On the other hand, when we shuffle identity-unrelated features, face color remains unchanged, while background of two images is swapped. This suggests that 1) identity-related features encode face color information which is one of the crucial factors for discriminating persons when they change their outfits, and 2) identity-unrelated ones hold background information.

### 4.3.3 Disentangled Features

Fig. 14a show examples for the identity-shuffled generation. To synthesize new images, we use identity-related features extracted from source images and identity-unrelated features extracted from target ones. We can observe that synthesized images share the same pose and background with target images, while clothing color changes according to the source images, suggesting that identity-unrelated and -related features encode pose/background and clothing color information, respectively.

To further demonstrate the effectiveness of identity-related features, we perform an experiment with attribute labels [84], which are useful for describing person identities, for Market-1501 [28], including gender, long/short hair,

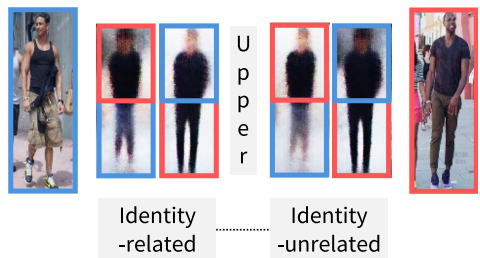

Fig. 13. An example of generated images using a part-level identity shuffling technique on Celeb-reID. We generate person images by shuffling identity-related/-unrelated features for upper body parts between two images. We shuffle either identity-related or -unrelated features while fixing the other. (Best viewed in color.)

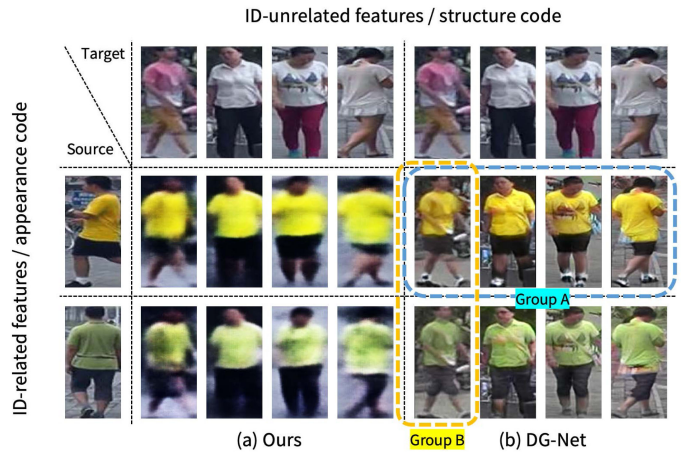

Fig. 14. An example of generated images by (a) ours and (b) DG-Net. We generate person images with identity-related/appearance features of the source images and identity-unrelated/structure ones of the target images. (Best viewed in color.)

TABLE 6
Quantitative Results for Classifying Person Attributes
on the Gallery Set of Market-1501

|  | ID-related | ID-unrelated |
|---|---|---|
| Gender | 96.88 | 51.56 |
| Hair length | 90.92 | 55.60 |
| Age | 72.69 | 58.98 |
| Pant/Dress | 78.42 | 56.58 |
| Color of upper-body clothing | 81.95 | 58.90 |
| Color of lower-body clothing | 75.07 | 58.62 |
| Length of upper-body clothing | 75.00 | 59.21 |
| Length of lower-body clothing | 94.60 | 54.95 |
| Carrying handbag/backpack | 74.10 | 57.58 |

*We report classification accuracies (%).*

clothing color, and sleeve length. Specifically, following the linear classification protocol of recent self-supervised learning methods, e.g., MoCo [85], we freeze identity-related/-unrelated features of our model, e.g., IS-GAN$_{KL}$, and train a linear classifier (a single fully-connected layer followed by a sigmoid function) for each attribute using a binary cross-entropy loss to see whether our features encode such attribute information. The intuition behind this experiment is that if features encode information of certain attribute, they could be linearly separable w.r.t labels of the attribute. For example, if features contain gender information, features extracted from images containing male persons are closely embedded, forming a group, and the same for those for the female. Specifically, we split the gallery set of Market-1501 into train/validation samples with a 3:1 ratio, and train the linear classifier for 10 epochs over the train set with the learning rate of 1e-2, where the learning rate decays by 10 every 3 epochs. We set the size of mini-batches to 64. We aggregate similar attributes into a group, and report a mean accuracy of each group on the validation subset in Table 6. We can observe that the accuracies remain around 50%, equal to the probability of random selections, when using identity-unrelated features. On the other hand, using identity-related features, we can obtain much higher accuracies, e.g., 96.88%, 94.60%, and 90.92% for gender, length of lower-body clothing, and hair length, respectively. These demonstrate that our model successfully disentangles person images into identity-related/-unrelated features, and identity-related ones capture important cues for specifying persons such as gender, length of lower-body clothing, and hair length.

### 4.3.4 Comparison With DG-Net [31]

DG-Net [31] aims at coupling data augmentation and feature learning for reID. To this end, following AdaIN [43], it transforms the structure feature using the appearance feature by simply scaling and shifting the normalized structure one to synthesize new images. Although AdaIN has shown the effectiveness to alter the color and/or texture of an image, it does not change semantic contents. See Fig. 14b for example. This suggests that 1) DG-Net does change the clothing color of target images in accordance with that of source images, and 2) physical characteristics, e.g., gender, of target images remain unchanged. In fact, generated images using the same appearance feature, e.g., Group A in Fig. 14b, are regarded as containing persons of the same identity when training the generator for the data augmentation (Eq. (6) in DG-Net [31]). For feature learning, on the other hand, generated images that share the same structural feature, e.g., Group B in Fig. 14b, are considered to have the same identity (Eq. (9) in DG-Net [31]), indicating that DG-Net fails to synthesize identity-shuffled images. We perform the linear classification protocol with the appearance/structure features of DG-Net. For the gender label, we obtain 57.26% and 56.05% accuracies after training, based on the frozen appearance and structure features, respectively. For other attributes, linear classifiers consistently output 0 or 1 regardless of input features, after few iterations. These indicate appearance and structure features are not linearly separable w.r.t person attributes, and they may not encode such information. Our model, on the other hand, synthesizes person images using the addition of identity-related and -unrelated features rather than simply changing the distribution of the features. This makes our generator reconstruct the original image from a single vector, where spatial information of the image is lost, resulting in reconstructed images to be blurry with fewer scene details (Fig. 14a). However, our framework does not restrict information identity-related features could encode, e.g., color or style, and encourages them to contain other cues, e.g., gender, through identity/part-level shuffling terms. As a result, our models outperform DG-Net for all benchmarks as shown in Table 1.

## 5 CONCLUSION

We have presented a novel framework learning disentangled representations for robust person reID. In particular, we have proposed a feature disentanglement method using an identity shuffling technique, which allows to factorize images into identity-related/-unrelated features without any auxiliary supervisory signals. To facilitate the disentanglement process, we have exploited a KL divergence loss, widely adopted for feature regularization, and also introduced a decorrelation loss making identity-related/-unrelated features mutually exclusive explicitly. We achieve a new state of the art on both short-term and long-term reID tasks in terms of rank-1 accuracy and mAP.

## REFERENCES

[1] M. Koestinger, M. Hirzer, P. Wohlhart, P. M. Roth, and H. Bischof, "Large scale metric learning from equivalence constraints," in *Proc. IEEE Conf. Comput. Vis. Pattern Recognit.*, 2012, pp. 2288–2295.

[2] S. Liao, Y. Hu, X. Zhu, and S. Z. Li, "Person re-identification by local maximal occurrence representation and metric learning," in *Proc. IEEE Conf. Comput. Vis. Pattern Recognit.*, 2015, pp. 2197–2206.

[3] S. Liao and S. Z. Li, "Efficient PSD constrained asymmetric metric learning for person re-identification," in *Proc. IEEE Int. Conf. Comput. Vis.*, 2015, pp. 3685–3693.

[4] A. Hermans, L. Beyer, and B. Leibe, "In defense of the triplet loss for person re-identification," 2017, *arXiv:1703.07737*.

[5] W. Chen, X. Chen, J. Zhang, and K. Huang, "Beyond triplet loss: A deep quadruplet network for person re-identification," in *Proc. IEEE Conf. Comput. Vis. Pattern Recognit.*, 2017, pp. 403–412.

[6] Y. Shen, H. Li, S. Yi, D. Chen, and X. Wang, "Person re-identification with deep similarity-guided graph neural network," in *Proc. Eur. Conf. Comput. Vis.*, 2018, pp. 486–504.

[7] M. M. Kalayeh, E. Basaran, M. Gökmen, M. E. Kamasak, and M. Shah, "Human semantic parsing for person re-identification," in *Proc. IEEE Conf. Comput. Vis. Pattern Recognit.*, 2018, pp. 1062–1071.

[8] Y. Shen, T. Xiao, H. Li, S. Yi, and X. Wang, "End-to-end deep kronecker-product matching for person re-identification," in *Proc. IEEE Conf. Comput. Vis. Pattern Recognit.*, 2018, pp. 6886–6895.

[9] H. Zhao et al., "Spindle Net: Person re-identification with human body region guided feature decomposition and fusion," in Proc. IEEE Conf. Comput. Vis. Pattern Recognit., 2017, pp. 1077–1085.

[10] C. Su, J. Li, S. Zhang, J. Xing, W. Gao, and Q. Tian, "Pose-driven deep convolutional model for person re-identification," in Proc. IEEE Int. Conf. Comput. Vis., 2017, pp. 3960–3969.

[11] Y. Sun, L. Zheng, Y. Yang, Q. Tian, and S. Wang, "Beyond part models: Person retrieval with refined part pooling (and a strong convolutional baseline)," in Proc. Eur. Conf. Comput. Vis., 2018, pp. 480–496.

[12] G. Wang, Y. Yuan, X. Chen, J. Li, and X. Zhou, "Learning discriminative features with multiple granularities for person re-identification," in Proc. 26th ACM Int. Conf. Multimedia, 2018, pp. 274–282.

[13] L. Zhao, X. Li, Y. Zhuang, and J. Wang, "Deeply-learned part-aligned representations for person re-identification," in Proc. IEEE Int. Conf. Comput. Vis., 2017, pp. 3219–3228.

[14] X. Liu et al., "HydraPlus-Net: Attentive deep features for pedestrian analysis," in Proc. IEEE Int. Conf. Comput. Vis., 2017, pp. 350–359.

[15] W. Li, X. Zhu, and S. Gong, "Harmonious attention network for person re-identification," in Proc. IEEE Conf. Comput. Vis. Pattern Recognit., 2018, pp. 2285–2294.

[16] D. Li, X. Chen, Z. Zhang, and K. Huang, "Learning deep context-aware features over body and latent parts for person re-identification," in Proc. IEEE Conf. Comput. Vis. Pattern Recognit., 2017, pp. 384–393.

[17] X. Zhang et al., "AlignedReID: Surpassing human-level performance in person re-identification," 2017, arXiv:1711.08184.

[18] K. He, X. Zhang, S. Ren, and J. Sun, "Deep residual learning for image recognition," in Proc. IEEE Conf. Comput. Vis. Pattern Recognit., 2016, pp. 770–778.

[19] A. Krizhevsky, I. Sutskever, and G. E. Hinton, "ImageNet classification with deep convolutional neural networks," in Proc. 25th Int. Conf. Neural Inf. Process. Syst., 2012, pp. 1097–1105.

[20] I. Goodfellow et al., "Generative adversarial nets," in Proc. 27th Int. Conf. Neural Inf. Process. Syst., 2014, pp. 2672–2680.

[21] X. Qian et al., "Pose-normalized image generation for person re-identification," in Proc. Eur. Conf. Comput. Vis., 2018, pp. 650–667.

[22] Y. Li, T. Zhang, L. Duan, and C. Xu, "A unified generative adversarial framework for image generation and person re-identification," in Proc. 26th ACM Int. Conf. Multimedia, 2018, pp. 163–172.

[23] M. F. Mathieu, J. J. Zhao, J. Zhao, A. Ramesh, P. Sprechmann, and Y. LeCun, "Disentangling factors of variation in deep representation using adversarial training," in Proc. 30th Int. Conf. Neural Inf. Process. Syst., 2016, pp. 5040–5048.

[24] J. Bao, D. Chen, F. Wen, H. Li, and G. Hua, "Towards open-set identity preserving face synthesis," in Proc. IEEE Conf. Comput. Vis. Pattern Recognit., 2018, pp. 6713–6722.

[25] H.-Y. Lee, H.-Y. Tseng, J.-B. Huang, M. Singh, and M.-H. Yang, "Diverse image-to-image translation via disentangled representations," in Proc. Eur. Conf. Comput. Vis., 2018, pp. 35–51.

[26] B. Lu, J.-C. Chen, and R. Chellappa, "Unsupervised domain-specific deblurring via disentangled representations," in Proc. IEEE Conf. Comput. Vis. Pattern Recognit., 2019, pp. 10 225–10 234.

[27] C. Eom and B. Ham, "Learning disentangled representation for robust person re-identification," in Proc. Int. Conf. Neural Inf. Process. Syst., 2019, pp. 5297–5308.

[28] L. Zheng, L. Shen, L. Tian, S. Wang, J. Wang, and Q. Tian, "Scalable person re-identification: A benchmark," in Proc. IEEE Int. Conf. Comput. Vis., 2015, pp. 1116–1124.

[29] W. Li, R. Zhao, T. Xiao, and X. Wang, "DeepReID: Deep filter pairing neural network for person re-identification," in Proc. IEEE Conf. Comput. Vis. Pattern Recognit., 2014, pp. 152–159.

[30] Z. Zheng, L. Zheng, and Y. Yang, "Unlabeled samples generated by GAN improve the person re-identification baseline in vitro," in Proc. IEEE Int. Conf. Comput. Vis., 2017, pp. 3754–3762.

[31] Z. Zheng, X. Yang, Z. Yu, L. Zheng, Y. Yang, and J. Kautz, "Joint discriminative and generative learning for person re-identification," in Proc. IEEE Conf. Comput. Vis. Pattern Recognit., 2019, pp. 2138–2147.

[32] Y. Suh, J. Wang, S. Tang, T. Mei, and K. Mu Lee, "Part-aligned bilinear representations for person re-identification," in Proc. Eur. Conf. Comput. Vis., 2018, pp. 402–419.

[33] J. Guo, Y. Yuan, L. Huang, C. Zhang, J.-G. Yao, and K. Han, "Beyond human parts: Dual part-aligned representations for person re-identification," in Proc. IEEE Int. Conf. Comput. Vis., 2019, pp. 3642–3651.

[34] J. Miao, Y. Wu, P. Liu, Y. Ding, and Y. Yang, "Pose-guided feature alignment for occluded person re-identification," in Proc. IEEE Int. Conf. Comput. Vis., 2019, pp. 542–551.

[35] B. Chen, W. Deng, and J. Hu, "Mixed high-order attention network for person re-identification," in Proc. IEEE Int. Conf. Comput. Vis., 2019, pp. 371–381.

[36] T. Chen et al., "ABD-net: Attentive but diverse person re-identification," in Proc. IEEE Int. Conf. Comput. Vis., 2019, pp. 8351–8361.

[37] G. Chen, C. Lin, L. Ren, J. Lu, and J. Zhou, "Self-critical attention learning for person re-identification," in Proc. IEEE Int. Conf. Comput. Vis., 2019, pp. 9637–9646.

[38] L. Wei, S. Zhang, W. Gao, and Q. Tian, "Person transfer GAN to bridge domain gap for person re-identification," in Proc. IEEE Conf. Comput. Vis. Pattern Recognit., 2018, pp. 79–88.

[39] J. Liu, "Identity preserving generative adversarial network for cross-domain person re-identification," IEEE Access, vol. 7, pp. 114021–114032, 2019.

[40] Y. Ge et al., "FD-GAN: Pose-guided feature distilling GAN for robust person re-identification," in Proc. 32nd Int. Conf. Neural Inf. Process. Syst., 2018, pp. 1222–1233.

[41] J.-Y. Zhu, T. Park, P. Isola, and A. A. Efros, "Unpaired image-to-image translation using cycle-consistent adversarial networks," in Proc. IEEE Int. Conf. Comput. Vis., 2017, pp. 2223–2232.

[42] Y. Choi, M. Choi, M. Kim, J.-W. Ha, S. Kim, and J. Choo, "StarGAN: Unified generative adversarial networks for multi-domain image-to-image translation," in Proc. IEEE Conf. Comput. Vis. Pattern Recognit., 2018, pp. 8789–8797.

[43] X. Huang and S. Belongie, "Arbitrary style transfer in real-time with adaptive instance normalization," in Proc. IEEE Int. Conf. Comput. Vis., 2017, pp. 1501–1510.

[44] Y. Zou, X. Yang, Z. Yu, B. Kumar, and J. Kautz, "Joint disentangling and adaptation for cross-domain person re-identification," in Proc. Eur. Conf. Comput. Vis., 2020, pp. 87–104.

[45] K. Zhou, Y. Yang, A. Cavallaro, and T. Xiang, "Omni-scale feature learning for person re-identification," in Proc. IEEE Int. Conf. Comput. Vis., 2019, pp. 3702–3712.

[46] L. Zheng, Y. Yang, and A. G. Hauptmann, "Person re-identification: Past, present and future," 2016, arXiv:1610.02984.

[47] X. Chang, T. M. Hospedales, and T. Xiang, "Multi-level factorisation net for person re-identification," in Proc. IEEE Conf. Comput. Vis. Pattern Recognit., 2018, pp. 2109–2118.

[48] H. Huang, J. Xu, Q. Wu, Y. Zhong, P. Zhang, and Z. Zhang, "Beyond scalar neuron: Adopting vector-neuron capsules for long-term person re-identification," IEEE Trans. Circuits Syst. Video Technol., vol. 30, no. 10, pp. 3459–3471, Oct. 2020.

[49] S. Sabour, N. Frosst, and G. E. Hinton, "Dynamic routing between capsules," in Proc. 31st Int. Conf. Neural Inf. Process. Syst., 2017, pp. 3856–3866.

[50] S. Yu, S. Li, D. Chen, R. Zhao, J. Yan, and Y. Qiao, "COCAS: A large-scale clothes changing person dataset for re-identification," in Proc. IEEE/CVF Conf. Comput. Vis. Pattern Recognit., 2020, pp. 3397–3406.

[51] Y.-J. Li, Z. Luo, X. Weng, and K. M. Kitani, "Learning shape representations for clothing variations in person re-identification," 2020, arXiv:2003.07340.

[52] Y. Liu, F. Wei, J. Shao, L. Sheng, J. Yan, and X. Wang, "Exploring disentangled feature representation beyond face identification," in Proc. IEEE Conf. Comput. Vis. Pattern Recognit., 2018, pp. 2080–2089.

[53] X. Huang, M.-Y. Liu, S. Belongie, and J. Kautz, "Multimodal unsupervised image-to-image translation," in Proc. Eur. Conf. Comput. Vis., 2018, pp. 172–189.

[54] L. Tran, X. Yin, and X. Liu, "Disentangled representation learning GAN for pose-invariant face recognition," in Proc. IEEE Conf. Comput. Vis. Pattern Recognit., 2017, pp. 1415–1424.

[55] I. Gel'fand and A. Yaglom, "Calculation of the amount of information about a random function contained in another such function," Amer. Math. Soc. Translations Series, vol. 2, no. 12, pp. 191–198, 1959.

[56] B. Cheung, J. A. Livezey, A. K. Bansal, and B. A. Olshausen, "Discovering hidden factors of variation in deep networks," in Proc. Int. Conf. Learn. Representations Workshop, 2015.

[57] Y. LeCun, L. Bottou, Y. Bengio, and P. Haffner, "Gradient-based learning applied to document recognition," Proc. IEEE, vol. 86, no. 11, pp. 2278–2324, Nov. 1998.

[58] L. Kristoufek, "Measuring correlations between non-stationary series with DCCA coefficient," Physica A: Statist. Mechanics Appl., vol. 402, pp. 291–298, 2014.

[59] A. Odena, C. Olah, and J. Shlens, "Conditional image synthesis with auxiliary classifier GANs," in *Proc. 34th Int. Conf. Mach. Learn.*, 2017, pp. 2642–2651.

[60] S. Ioffe and C. Szegedy, "Batch normalization: Accelerating deep network training by reducing internal covariate shift," in *Proc. 32nd Int. Conf. Mach. Learn.*, 2015, pp. 448–456.

[61] A. L. Maas, A. Y. Hannun, and A. Y. Ng, "Rectifier nonlinearities improve neural network acoustic models," in *Proc. Int. Conf. Mach. Learn.*, 2013, Art. no. 3.

[62] N. Srivastava, G. Hinton, A. Krizhevsky, I. Sutskever, and R. Salakhutdinov, "Dropout: A simple way to prevent neural networks from overfitting," *J. Mach. Learn. Res.*, vol. 15, no. 1, pp. 1929–1958, 2014.

[63] D. Ulyanov, A. Vedaldi, and V. Lempitsky, "Instance normalization: The missing ingredient for fast stylization," 2016, *arXiv:1607.08022*.

[64] P. Isola, J.-Y. Zhu, T. Zhou, and A. A. Efros, "Image-to-image translation with conditional adversarial networks," in *Proc. IEEE Conf. Comput. Vis. Pattern Recognit.*, 2017, pp. 1125–1134.

[65] D. P. Kingma and M. Welling, "Auto-encoding variational bayes," in *Proc. Int. Conf. Learn. Representations*, 2014.

[66] Z. Zhong, L. Zheng, D. Cao, and S. Li, "Re-ranking person re-identification with k-reciprocal encoding," in *Proc. IEEE Conf. Comput. Vis. Pattern Recognit.*, 2017, pp. 1318–1327.

[67] E. Ristani, F. Solera, R. Zou, R. Cucchiara, and C. Tomasi, "Performance measures and a data set for multi-target, multi-camera tracking," in *Proc. Eur. Conf. Comput. Vis.*, 2016, pp. 17–35.

[68] D. P. Kingma and J. Ba, "Adam: A method for stochastic optimization," in *Proc. Int. Conf. Learn. Representations*, 2015.

[69] Z. Zhong, L. Zheng, G. Kang, S. Li, and Y. Yang, "Random erasing data augmentation," in *Proc. 34th AAAI Conf. Artif. Intell.*, 2020, pp. 13 001–13 008.

[70] H. Luo, Y. Gu, X. Liao, S. Lai, and W. Jiang, "Bag of tricks and a strong baseline for deep person re-identification," in *Proc. IEEE/CVF Conf. Comput. Vis. Pattern Recognit. Workshop*, 2019, pp. 1487–1495.

[71] R. Quispe and H. Pedrini, "Top-DB-net: Top dropblock for activation enhancement in person re-identification," in *Proc. 25th Int. Conf. Pattern Recognit.*, 2021, pp. 2980–2987.

[72] X. Ni, L. Fang, and H. Huttunen, "Adaptive L2 regularization in person re-identification," in *Proc. 25th Int. Conf. Pattern Recognit.*, 2021, pp. 9601–9607.

[73] Y. Sun, L. Zheng, W. Deng, and S. Wang, "SVDNet for pedestrian retrieval," in *Proc. IEEE Int. Conf. Comput. Vis.*, 2017, pp. 3800–3808.

[74] Y. Wang et al., "Resource aware person re-identification across multiple resolutions," in *Proc. IEEE Conf. Comput. Vis. Pattern Recognit.*, 2018, pp. 8042–8051.

[75] N. Martinel, G. Luca Foresti, and C. Micheloni, "Aggregating deep pyramidal representations for person re-identification," in *Proc. IEEE/CVF Conf. Comput. Vis. Pattern Recognit. Workshop*, 2019, pp. 1544–1554.

[76] C.-P. Tay, S. Roy, and K.-H. Yap, "AANet: Attribute attention network for person re-identifications," in *Proc. IEEE Conf. Comput. Vis. Pattern Recognit.*, 2019, pp. 7134–7143.

[77] Y. Fu et al., "Horizontal pyramid matching for person re-identification," in *Proc. 33rd AAAI Conf. Artif. Intell.*, 2019, pp. 8295–8302.

[78] M. Zheng, S. Karanam, Z. Wu, and R. J. Radke, "Re-identification with consistent attentive siamese networks," in *Proc. IEEE Conf. Comput. Vis. Pattern Recognit.*, 2019, pp. 5735–5744.

[79] S. Zhao et al., "Do not disturb me: Person re-identification under the interference of other pedestrians," in *Proc. Eur. Conf. Comput. Vis.*, 2020, pp. 647–663.

[80] F. Zheng et al., "Pyramidal person re-identification via multi-loss dynamic training," in *Proc. IEEE Conf. Comput. Vis. Pattern Recognit.*, 2019, pp. 8514–8522.

[81] Y. Sun et al., "Circle loss: A unified perspective of pair similarity optimization," in *Proc. IEEE Conf. Comput. Vis. Pattern Recognit.*, 2020, pp. 6398–6407.

[82] X. Liang, K. Gong, X. Shen, and L. Lin, "Look into person: Joint body parsing & pose estimation network and a new benchmark," *IEEE Trans. Pattern Anal. Mach. Intell.*, vol. 41, no. 4, pp. 871–885, Apr. 2019.

[83] L. v. d. Maaten and G. Hinton, "Visualizing data using t-SNE," *J. Mach. Learn. Res.*, vol. 9, no. Nov., pp. 2579–2605, 2008.

[84] Y. Lin et al., "Improving person re-identification by attribute and identity learning," *Pattern Recognit.*, vol. 95, pp. 151–161, 2019.

[85] K. He, H. Fan, Y. Wu, S. Xie, and R. Girshick, "Momentum contrast for unsupervised visual representation learning," in *Proc. IEEE Conf. Comput. Vis. Pattern Recognit.*, 2020, pp. 9729–9738.

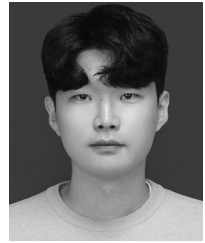

**Chanho Eom** received the BS degree in electrical and electronic engineering from Yonsei University, Seoul, South Korea, in 2017, where he is currently working toward the joint MS and PhD degrees in electrical and electronic engineering. His current research interests include computer vision and machine learning, in particular, re-identification and video analysis, both in theory and applications.

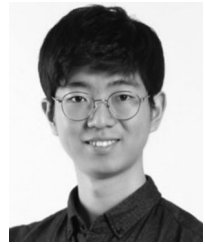

**Wonkyung Lee** received the BS degree in electrical and electronic engineering from Yonsei University, Seoul, South Korea, in 2019. He is currently working toward the PhD degree in Electrical and Electronic Engineering Department, Yonsei University, South Korea. His research interests include building efficient deep learning models for visual tasks. He is also interested in developing alternative self-attention mechanisms for visual recognition.

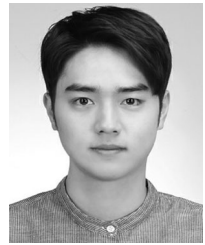

**Geon Lee** received the BS degree in electrical and electronic engineering from Yonsei University, Seoul, South Korea, in 2019. He is currently working toward the joint MS and PhD degrees from Yonsei University, South Korea. He is interested in solving vision tasks including object detection and semantic segmentation.

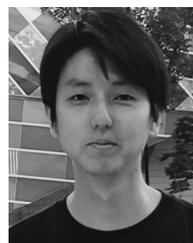

**Bumsub Ham** (Member, IEEE) received the BS and PhD degrees in electrical and electronic engineering from Yonsei University, Korea, in 2008 and 2013, respectively. He is currently an associate professor of electrical and electronic engineering at Yonsei University, Seoul, Korea. From 2014 to 2016, he was postdoctoral research fellow with Willow Team of INRIA Rocquencourt, École Normale Supérieure de Paris, and Centre National de la Recherche Scientifique. His research interests include computer vision, computational photography, and machine learning, in particular, regularization and matching, both in theory and applications.

▷ **For more information on this or any other computing topic, please visit our Digital Library at** www.computer.org/csdl.